%% file: main.tex
\documentclass{article} %
\usepackage{arxiv_preprint,times}

\usepackage{graphicx}
\usepackage{wrapfig}
\usepackage{booktabs}
\usepackage{makecell}
\usepackage{siunitx}
\usepackage{amsmath,amssymb}
\usepackage{subcaption}
\usepackage{agora_abbrv}
\usepackage{soul}
\usepackage{comment}

\usepackage{hyperref}
\usepackage{url}

\usepackage{xcolor}
\usepackage{soul}
\usepackage[framemethod=TikZ]{mdframed}
\sethlcolor{yellow}

\newif\ifhighlightchanges
\highlightchangesfalse

\soulregister{\eg}{0}
\soulregister{\ie}{0}
\soulregister{\cf}{0}
\soulregister{\etal}{0}
\soulregister{\etc}{0}
\soulregister{\vs}{0}

\ifhighlightchanges
  
  \newmdenv[hidealllines=true,
            backgroundcolor=yellow,
            innerleftmargin=2pt, innerrightmargin=2pt,
            innertopmargin=2pt,  innerbottommargin=2pt,
            skipabove=0pt, skipbelow=0pt]{yellowblock}
\else

\fi

\newif\ifteaserfig

\newlength{\qlabw}
\newcommand{\qrow}[2]{%
  \noindent\begin{minipage}[c]{\qlabw}\centering\rotatebox{90}{\scriptsize #1}\end{minipage}%
  \begin{minipage}[c]{\dimexpr\linewidth-\qlabw\relax}%
    \includegraphics[width=\linewidth]{#2}\end{minipage}\par\vspace{1.5pt}}
\newcommand{\qh}[2]{\makebox[\dimexpr(\linewidth-\qlabw)/#1\relax][c]{\scriptsize #2}}
\newcommand{\qhh}[2]{\makebox[\dimexpr(\linewidth-\qlabw)*2/#1\relax][c]{\scriptsize #2}}

\newcommand{\gr}[1]{\textcolor{gray}{#1}}

\preprintfinalcopy

\title{One Basis to Animate Them All: \\Gaussian Blendshape Distillation for\\ Real-Time Avatars}

\author{Ramazan Fazylov \\
Mohamed bin Zayed University of Artificial Intelligence \\
\texttt{ramazan.fazylov@mbzuai.ac.ae} \\
\And
Stamatis Lefkimmiatis \\
MWS AI \\
\texttt{s.lefkimmiatis@mts.ai} \\
\And
Ivan Laptev \\
Mohamed bin Zayed University of Artificial Intelligence \\
\texttt{ivan.laptev@mbzuai.ac.ae}
}

\hypersetup{
  pdftitle={One Basis to Animate Them All: Gaussian Blendshape Distillation for Real-Time Avatars},
  pdfauthor={Ramazan Fazylov, Stamatis Lefkimmiatis, Ivan Laptev}
}

\belowdisplayskip \abovedisplayskip
\makeatletter
\def\section{\@startsection {section}{1}{\z@}{-1.4ex plus -0.4ex minus -.2ex}{0.9ex plus 0.2ex minus 0.2ex}{\large\sc\raggedright}}
\def\subsection{\@startsection{subsection}{2}{\z@}{-1.2ex plus -0.4ex minus -.2ex}{0.5ex plus .2ex}{\normalsize\sc\raggedright}}
\def\subsubsection{\@startsection{subsubsection}{3}{\z@}{-1.0ex plus -0.4ex minus -.2ex}{0.3ex plus .2ex}{\normalsize\sc\raggedright}}
\def\paragraph{\@startsection{paragraph}{4}{\z@}{0.8ex plus 0.4ex minus .2ex}{-1em}{\normalsize\bf}}
\makeatother

\begin{document}

\maketitle

\input{sec/0_abstract}
\input{sec/1_intro}
\input{sec/2_related_work}
\input{sec/3_method}
\input{sec/4_experiments}
\input{sec/6_conclusion}

\bibliography{main}

\clearpage
\appendix
\section*{Appendix}
\input{sec/X_suppl}

\end{document}

%% file: sec/0_abstract.tex
\IfFileExists{figures/teaser_prefinal_v3.pdf}{\teaserfigtrue}{\teaserfigfalse}
\ifteaserfig
\begin{figure}[!htb]
\centering
\mbox{}\vspace{-.3cm}\\
\includegraphics[width=0.95\linewidth]{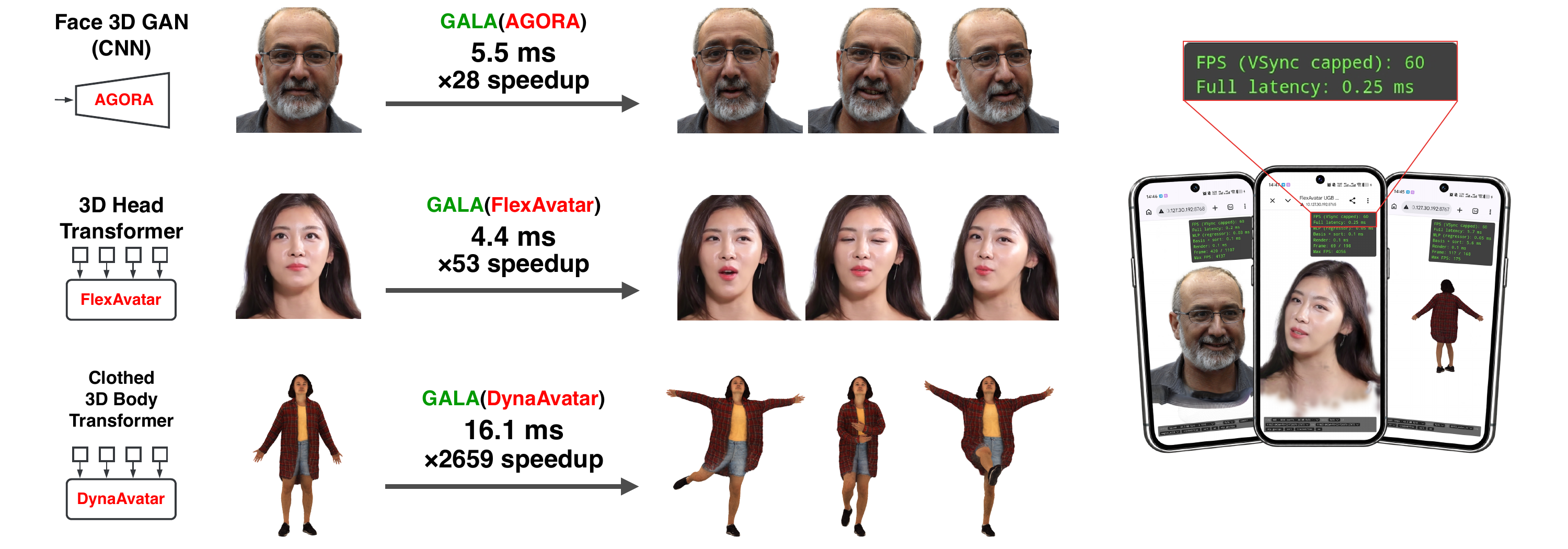}
\caption{
GALA enables efficient, high-fidelity animation across diverse Gaussian avatar models. By introducing shared blendshapes and a lightweight coefficient predictor, GALA accelerates CPU animation by up to three orders of magnitude and enables real-time deployment on mobile devices.
}
\label{fig:teaser}
\end{figure}
\fi

\begin{abstract}
3D Gaussian avatars support fast rendering, however, their real-time animation is often challenged by the costly neural inference. We address this bottleneck and show that the animation of pretrained avatar models can be closely approximated by a linear combination of identity-independent blendshapes. 
Building on this finding, we introduce \textbf{GALA} (\textbf{G}aussian \textbf{A}nimation via \textbf{L}inear \textbf{A}pproximation), a distillation method that replaces per-frame heavy neural decoding with a shallow coefficient predictor and a linear blend. 
To improve fidelity and reduce memory requirements, we propose to construct the basis using block-local PCA under a rendering-aware metric and a memory budget. Our method learns a shallow MLP network to predict blendshape coefficients and applies to various animation architectures without retraining original models. We validate GALA by accelerating the inference of three distinct avatar models for 3D animation of facial expressions and full-bodies with clothing dynamics. Across these models, our distillation generalizes to held-out identities and reduces CPU animation cost by up to three orders of magnitude while preserving most of the rendering quality. Excellent results of our method confirm the shared linear structure of learned avatar representations and enable highly efficient and accurate animation at frame rates reaching up to 60fps on mobile devices. {\em Project page: \url{https://ramazan793.github.io/gala/}}

\end{abstract}

%% file: sec/1_intro.tex
\section{Introduction}
\label{sec:intro}

Recent advances in neural avatars have substantially improved the generation, reconstruction, and animation of photorealistic digital humans. Representations based on 3D Gaussians are particularly attractive because their explicit primitives support efficient rendering~\citep{3dgs}, and they underlie models for head generation, reconstruction from a single image, and full-body animation with clothing dynamics~\citep{gaia,agora,flexavatar,dynavatar}. However, efficient rendering does not imply efficient animation. Many models evaluate a substantial neural network at every frame to update Gaussian attributes from expression or pose parameters, and this per-frame computation, which we call the model's \emph{animation path}, can dominate inference cost and limits deployment on CPUs and mobile devices.

\begin{table}[t]
\centering
\caption{
Comparison of avatar animation representations. Our method extracts an identity-shared blendshape basis from a pretrained nonlinear avatar model, complementing approaches based on predefined rigs or per-subject fitting. The extracted basis closely approximates the model's learned deformations and supports efficient linear blending for unseen identities.
}
\label{tab:positioning}
\footnotesize
\vspace{2pt}
\setlength{\tabcolsep}{3pt}
\begin{tabular}{@{}lccc l@{}}
\toprule
& \makecell{Photo-\\real.} & \makecell{Identity-\\shared} & \makecell{Linear at\\run time} & Where the basis comes from \\
\midrule
3D morphable models \citep{FLAME}
    & $\times$   & \checkmark & \checkmark & hand-built from scans \\
Per-person blendshapes \citep{zielonka2025gem}
    & \checkmark & $\times$   & \checkmark & fitted per subject \\
Fixed-rig avatars \citep{lam}
    & \checkmark & \checkmark & \checkmark & chosen up front \\
Animatable avatar models \citep{gaia}
    & \checkmark & --         & $\times$   & none (a network) \\
\midrule
GALA (Ours)
    & \checkmark & \checkmark & \checkmark & \textbf{discovered in a trained model} \\
\bottomrule
\end{tabular}
\end{table}

The cost of neural animation raises the question of how much of the computation performed by a pretrained avatar model is necessary at inference time. The complexity of its nonlinear decoder does not establish that its outputs require an equally complex representation, and deformations across identities may share a compact linear structure even when the architecture and the training objective do not impose one. In this work, we show that the animation outputs of diverse pretrained Gaussian avatar models can be closely approximated by an identity-specific neutral avatar and a linear combination of identity-independent {\em blendshapes}, which we obtain by analyzing Gaussian attribute residuals relative to the neutral state across identities and driving signals. Crucially, a basis extracted once from a model also captures the animation of identities excluded from basis construction, since identity-dependent motion is retained through the blendshape coefficients while the deformation basis is shared.

Linear representations have a long history in face and body modeling~\citep{blanz1999morphable,FLAME}. More recently, Gaussian blendshape methods have demonstrated efficient photorealistic animation with bases fitted to individual subjects~\citep{ma2024gaussianblendshapes,zielonka2025gem}, while systems such as LAM adopt a predefined animation rig~\citep{lam}. Our work investigates a different setting, extracting a reusable basis from a trained model that represents many identities (Table~\ref{tab:positioning}). The basis approximates deformations learned by the original network and can be reused for new subjects without reconstructing it, where identity independence refers to sharing within one pretrained model.

Building on this finding, we introduce a Gaussian blendshape distillation strategy, which we call \textbf{GALA} (\textbf{G}aussian \textbf{A}nimation via \textbf{L}inear \textbf{A}pproximation). Our method constructs separate bases for Gaussian attribute groups and spatial regions using block-local principal component analysis under a rendering-aware metric, which prioritizes deformations that matter for visual fidelity, and allocates components within a memory budget according to their estimated image-space error reduction per stored byte. A shallow coefficient predictor, conditioned on identity and driving signals, then learns to control the extracted basis, so that animation requires only coefficient prediction and a linear blend added to the cached neutral avatar. The procedure requires no retraining of the original model.

We apply our approach to three existing models with distinct architectures and training objectives, the \emph{hosts} of our distillation: AGORA~\citep{agora} for head generation, FlexAvatar~\citep{flexavatar} for one-shot head reconstruction, and DynaAvatar~\citep{dynavatar} for full-body reconstruction with clothing dynamics. 
GALA closely approaches animation accuracy of approximated host models and consistently outperforms other baselines on respective benchmarks.
Crucially, our method increases CPU animation speed of host models by up to {\em three orders of magnitude} and enables on-device photorealistic 3D animation at 60~fps. 
Figure~\ref{fig:teaser} illustrates qualitative results of our method.
In summary, our contributions are threefold:
\vspace{-.2cm}
\begin{enumerate}
    \item {\bfseries Identity-independent linearization of neural animation.} We demonstrate that animation produced by complex pretrained Gaussian avatar networks can be closely approximated using a shared linear blendshape basis that generalizes to identities unseen during extraction.
    \vspace{-.5cm}
    \item {\bfseries A method for basis construction and animation control.} We introduce GALA, combining rendering-aware, block-local basis construction under a memory budget with a shallow network that predicts blendshape coefficients from the identity and the driving signals.
    \vspace{-.1cm}
    \item {\bfseries Extensive experimental validation.} We apply our method to three existing models for head and full-body animation and demonstrate significant improvements when compared to the state of the art and compact neural baselines.
\end{enumerate}

%% file: sec/2_related_work.tex
\section{Related Work}
\label{sec:related_work}

\noindent
\textbf{Animatable avatar models.}
Recent work on photorealistic avatars explores several approaches. 3D-aware GANs trained from unstructured images collections generate heads as triplanes~\citep{eg3d} or Gaussians~\citep{gghead, gsgan, cgsgan}, and their animatable versions condition generation on the parameters of morphable models through deformation fields or expression branches~\citep{Next3D, anifacegan, OmniAvatar, GNARF, gaia, agora}. 
One-shot reconstruction methods regress an animatable head from a single image, either by using 3D GANs during training~\citep{p4d, p4dv2, voodoo3d} or by direct supervised reconstruction from tracked video~\citep{gpavatar, gaga, lam, flexavatar}, and diffusion priors pursue the same goal with synthesised supervision~\citep{cap4d, mvp4d}. 
Per-subject methods optimise Gaussian avatars from videos of particular people~\citep{GaussianAvatars, NPGA, monogausavatar}. Work on full-body animation follows similar approaches. 
One-shot full-body reconstruction methods~\citep{lhm, idol, persona} articulate Gaussians by SMPL-X skinning~\citep{smplx}, while DynaAvatar~\citep{dynavatar} adds pose-dependent clothing dynamics using a transformer conditioned on pose history. 
We apply our approach to two recent host models for head animation AGORA~\citep{agora}, FlexAvatar~\citep{flexavatar} and one host model for full-body animation DynaAvatar~\citep{dynavatar}, and demonstrate significant improvement of their inference speed.

\smallskip\noindent
\textbf{Linear face and body models.}
Linear representations have a long history in modeling human shape and appearance. Eigenfaces represent facial images in a low-dimensional PCA subspace~\citep{sirovich1987, turk1991eigenfaces}, while active appearance models capture variations in shape and texture~\citep{cootes2001aam}. Three-dimensional morphable models extend this principle to facial geometry and appearance learned from scans~\citep{blanz1999morphable}.
For animation, blendshape models express deformations as weighted combinations of offsets from a neutral shape~\citep{parke1972, lewis2014blendshapes}, and multilinear models disentangle identity and expression variation~\citep{vlasic2005multilinear,cao2014facewarehouse}. 
Modern parametric models, including FLAME~\citep{FLAME} and SMPL-X~\citep{smplx}, combine learned shape and corrective blendshapes with articulated skinning to represent expressive faces and bodies; see~\citet{egger2020morphable} for a review of 3D morphable face models.
These representations provide the driving parameters for the avatar models considered in our work. We show that the resulting neural animation outputs themselves admit a compact linear approximation: an identity-shared blendshape basis over Gaussian attributes, extracted from a pretrained network without direct fitting to scans.

\smallskip\noindent
\textbf{Photorealistic blendshape models.}
Recent work extends blendshape representations to photorealistic Gaussian avatars. Methods such as 3D Gaussian Blendshapes~\citep{ma2024gaussianblendshapes}, HeadGaS~\citep{dhamo2024headgas}, Gaussian D\'ej\`a-vu~\citep{yan2025gaussiandejavu}, RGBAvatar~\citep{rgbavatar}, GEM~\citep{zielonka2025gem} and WebAvatar~\citep{webavatar} learn {\em subject-specific} bases over Gaussian attributes or latent features from monocular or multi-view observations. 
RGBAvatar predicts blending coefficients from tracked expression parameters, while GEM extracts a basis from a trained subject-specific avatar network and predicts its coefficients from images. 
Similarly, SqueezeMe~\citep{squeezeme} distils per-subject full-body Gaussian avatars into linear corrective bases for efficient headset rendering. These methods establish the effectiveness of linear representations for individual avatars, but require a separate basis for each subject.
Fixed-rig approaches such as LAM~\citep{lam} instead support efficient animation through a predefined FLAME-based articulation model. 
Differently from previous work, our method extracts a shared identity-agnostic blendshape basis from a pretrained neural model. The basis captures deformations learned by that model and is reused unchanged for identities unseen during extraction, extending Gaussian blendshape distillation from individual subjects to models of diverse identities.

%% file: sec/3_method.tex
\section{Method}
\label{sec:method}

Our goal is to accelerate pretrained Gaussian avatar models ({\em host models}) by replacing their per-frame neural decoding with a fast inference based on linear blendshape representations. We first introduce a generic host model and its animation parameters in Section~\ref{subsec:prelim}. %
Section~\ref{subsec:ugb} then presents main parts of our method, namely the construction of identity-shared basis
and prediction of its coefficients from animation parameters.

\subsection{Preliminaries}
\label{subsec:prelim}

We consider avatar models that represent a subject using $n$ 3D Gaussians rendered by differentiable splatting~\citep{3dgs}. Each Gaussian is described by a $d$-dimensional attribute vector, which includes position, scale, rotation, opacity, and appearance, with additional skinning weights where required. 
Avatar animation is controlled by parameters derived from parametric face or body models such as FLAME~\citep{FLAME} or SMPL-X~\citep{smplx}. 
We denote the animation driving input by $\theta$ and its neutral setting by $\theta_0$. For models with temporal conditioning, $\theta$ includes the required motion history. An identity descriptor $w$, either supplied directly or extracted from an input image, remains fixed throughout an animation sequence. 
We define pretrained host models as $A(w,\theta) \rightarrow G \in\mathbb{R}^{nd}$, where the output concatenates the attribute vectors of the $n$ Gaussians.
Importantly, Gaussian indices in the model output $G$ must have consistent correspondence across variations in identity and animation parameters. This constraint is typically satisfied by existing models which initialize Gaussians on a shared UV grid or template-anchored point~\citep{flexavatar,gaia,agora,dynavatar}.

\subsection{Gaussian Blendshape Distillation}
\label{subsec:ugb}
\label{subsec:agora_m}

\begin{figure}[t]
\centering
\includegraphics[width=\linewidth]{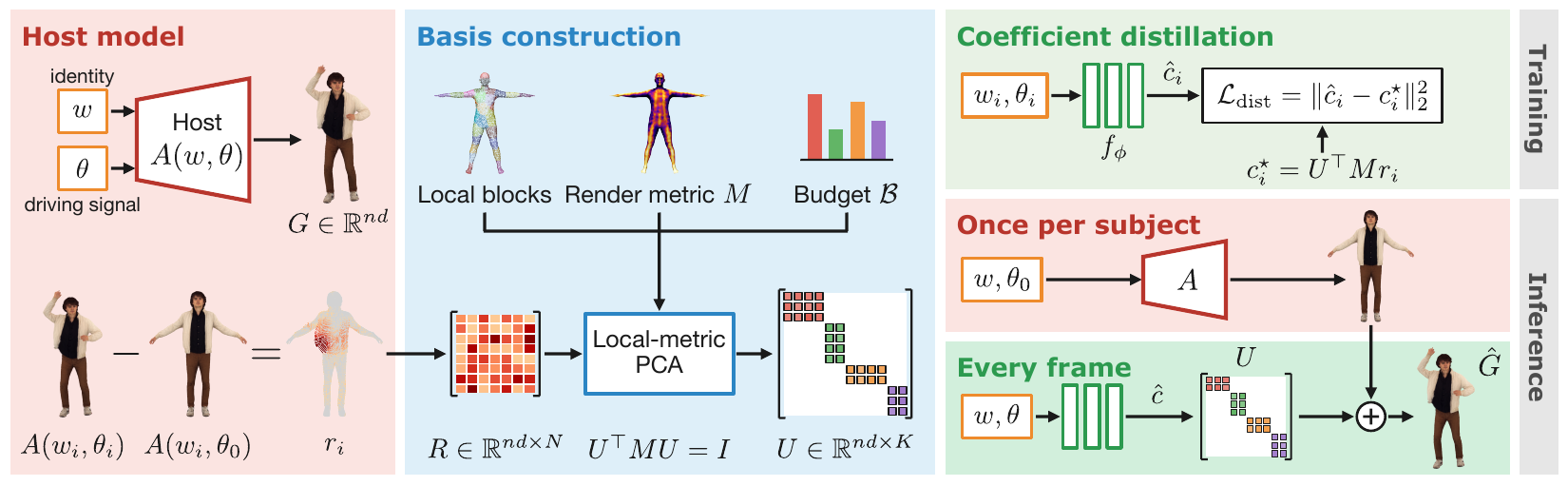}
\caption{Overview of GALA. Residuals $r_i$ of the host model $A$ with respect to the neutral avatar $A(w_i,\theta_0)$ (Section~\ref{subsubsec:simplepca}) are decomposed by a spatially local, rendering-aware PCA under a memory budget $\mathcal B$ into the block-diagonal basis $U$ (Section~\ref{subsubsec:improvedpca}). A coefficient network $f_\phi$ is trained to regress the projected coefficients $c^\star$ (Section~\ref{subsubsec:regression}). At inference, the neutral avatar is computed once per subject, and every frame costs one evaluation of a shallow network $f_\phi$ and a linear blend.}
\label{fig:model-overview}
\end{figure}

In this section we present our method that aims to approximate the costly neural animation of the original host model $G = A(w,\theta)$ by a simpler model 
\begin{equation}
    \hat{G} = A(w,\theta_0) + U c(w,\theta),
\label{eq:distillation}
\end{equation}
where $A(w,\theta_0)$ is the representation of a neutral avatar estimated only once per animation sequence, $U$ is a linear basis and $c$ is a vector of basis coefficients. The rest of this section explains our approach to construct the basis $U$ and to efficiently estimate its coefficients $c(w,\theta)$. Figure~\ref{fig:model-overview} gives an overview of our method.

\subsubsection{Simple PCA basis}
\label{subsubsec:simplepca}
Given a pretrained model $A(w,\theta)$ and the neutral pose $\theta_0$, we compute residuals $r$ for a large set of $N$ samples corresponding to identities $w_i$ and animation parameters $\theta_i$ as 
\begin{equation}
\label{eq:residual}
r_i=A(w_i,\theta_i)-A(w_i,\theta_0).
\end{equation}
We stack the residuals as columns of $R=[r_1,\dots,r_N]\in\mathbb{R}^{nd\times N}$. A rank-$K$ basis can then be obtained by solving
\begin{equation}
    \min_{U,\{c_i\}}
    \sum_{i=1}^{N}\|r_i-Uc_i\|_2^2,
    \qquad U^\top U=I.
\label{eq:simplepcabasis}
\end{equation}
By the Eckart--Young theorem~\citep{eckart1936approximation}, the solution is the truncated singular value decomposition $R\approx U\Sigma V^\top$, whose columns of $U$ are the top $K$ left singular vectors of $R$, with coefficients $c_i^\star=U^\top r_i$. We refer to this construction as uncentered PCA: residuals are measured relative to each identity’s neutral avatar, and no additional mean offset is introduced. Consequently, zero coefficients recover the neutral avatar exactly.

Because residuals from different identities are factorized jointly, the resulting basis describes shared animation variation. It can also be applied to identities excluded from extraction, provided that Gaussian indices retain the correspondence specified in Section~\ref{subsec:prelim}. However, this simple construction has three limitations. Attribute groups have different units and scales, global components incur storage costs even for localized deformations, and Euclidean attribute error does not directly reflect rendering quality. We address these limitations through the following modifications.

\subsubsection{Improved PCA basis}
\label{subsubsec:improvedpca}

\paragraph{Per-attribute and spatially local decomposition.} We first separate Gaussian attributes into groups $a$, each containing $d_a$ channels, such as position, scale, rotation, opacity, and appearance. Constructing a separate basis for each group avoids coupling attributes with different scales and allows their representation capacities to be selected independently.

Since animation deformations are often spatially localized, a global basis wastes memory storing components over the entire avatar when only a small region contributes to the motion. Following this intuition, we partition the Gaussians into $B$ spatial blocks using $k$-means such that the Gaussians within clusters share similar 3D locations.
For attribute group $a$ and block $j$, containing $n_j$ Gaussians, we construct a basis
\begin{equation}
    U_{a,j}\in\mathbb{R}^{n_jd_a\times K_{a,j}}   
\end{equation}
with $K_{a,j}$ components. The complete basis $U$ is formed by assembling these local bases in block-diagonal form after reordering the attributes. 
Each local component stores $n_jd_a$ values instead of $nd_a$ and encodes a specific spatial region of the avatar. To limit the size of the increased basis $U$, we reduce the number of components $K_{a,j}$ in $U_{a,j}$ allowing the memory budget to be concentrated on regions with significant variations and complex motion, see below. 

\paragraph{Rendering-aware PCA.} Standard PCA allocates representation capacity to directions with the largest variation in Gaussian attribute space. To prioritize deformations that affect visual quality, we replace the Euclidean reconstruction objective with a rendering-aware quadratic metric. Let $I(x)$ denote an image rendered from Gaussian attributes $x$, and let $J=\partial I/\partial x$. For a small attribute perturbation $\delta$,
\begin{equation}
    \|I(x+\delta)-I(x)\|_2^2
    \approx
    \delta^\top H\delta,
    \qquad H=J^\top J.
    \label{eq:attribute_perturbation}
\end{equation}
The full matrix $H$ is too large to construct. We therefore estimate only its per-Gaussian diagonal blocks within each attribute group by stochastic probing~\citep{hutchinson1989stochastic,bekas2007estimator} (Appendix~\ref{sec:supp_metric}). The estimate for attribute group $a$, normalized to unit mean trace per channel, is denoted as $\widehat H_a$. Applying the same procedure to the landmark positions instead of the image gives $\widehat H_a^{\mathrm{lmk}}$. We define the metric of attribute group $a$ as
\begin{equation}
M_a=\alpha I+\beta\widehat H_a+\gamma\widehat H_a^{\mathrm{lmk}}.
\label{eq:metric}
\end{equation}
The term $\beta\widehat H_a$ weights attribute errors by their effect on the rendered image, the ridge term $\alpha I$ with $\alpha>0$ retains sensitivity to directions that are weakly observed by the renderer, and the optional landmark term ($\gamma>0$) adds geometric supervision. Each local basis $U_{a,j}$ is then obtained by PCA of the residuals of block $j$ in the metric $M_{a,j}$ (Appendix~\ref{sec:supp_metric}). Setting $\beta=\gamma=0$ recovers the simple PCA of Section~\ref{subsubsec:simplepca} within each attribute group and block. Evaluating with projected coefficients $c_{a,j}^\star=U_{a,j}^\top M_{a,j}r_{a,j}$ isolates the approximation quality of the basis from errors introduced by coefficient prediction.

\paragraph{Memory-budgeted component allocation.} To maximize visual fidelity under a fixed memory budget, we select the numbers of components $K_{a,j}$ jointly across attribute groups and spatial blocks. Let $\lambda_{a,j,k}$ be the $k$th squared singular value in the PCA of block $j$ (Appendix~\ref{sec:supp_metric}). Since the PCA is computed in the metric $M_{a,j}$, $\lambda_{a,j,k}$ measures the variation of the residuals explained by component $k$ weighted by its effect on the rendered image. Undoing the per-group normalization of $\widehat H_a$ and the weight $\beta$ turns it into a value $v_{a,j,k}$ that is comparable across attribute groups, namely the estimated reduction of the squared image error of Eq.~\ref{eq:attribute_perturbation} summed over the samples (Appendix~\ref{sec:supp_metric}). Given a basis memory budget $\mathcal B$ and $b$ bytes per stored value, we maximize $\sum_{a,j}\sum_{k=1}^{K_{a,j}}v_{a,j,k}$ subject to $b\sum_{a,j}n_jd_aK_{a,j}\leq\mathcal B$, and solve this allocation greedily by adding components in the order of their value per byte, $v_{a,j,k}/(bn_jd_a)$, until the budget is exhausted. This allocates capacity according to estimated visual benefit and storage cost without manually specifying ranks for individual attributes or regions, uses the precomputed spectra and requires no additional rendering.

\subsubsection{Regressing blendshape coefficients}
\label{subsubsec:regression}

Projecting a residual of the host model onto the extracted basis provides accurate blendshape coefficients but requires evaluating the original animation network. To avoid this computation at inference time, we train a lightweight predictor
\begin{equation}
    \widehat c=f_\phi(w,\theta)\in\mathbb{R}^{K},
    \qquad K=\sum_{a,j}K_{a,j},
\end{equation}
which maps animation parameters $\theta$ and an identity descriptor $w$ to coefficients of the fixed basis. 
We optimize parameters $\phi$ of a shallow MLP network $f_\phi$, which we refer to as the {\em coefficient network}, by minimizing the distillation loss $\mathcal L_{\mathrm{dist}} = \|\widehat c - c^\star\|_2^2$ on the same samples $(w_i,\theta_i)$ we use for the basis construction. Appendix~\ref{sec:supp_coeffnet} details the architecture and training of the coefficient networks.
Identity conditioning allows the same expression or pose to produce subject-specific deformations despite sharing the basis across subjects.

%% file: sec/4_experiments.tex
\section{Experiments}
\label{sec:experiments}

\subsection{Experimental Setup}
\label{subsec:setup}

\paragraph{Host models.}
\label{subsec:hosts}
We apply our method to three host models with distinct architectures. AGORA~\citep{agora} generates heads with a GAN and animates canonical Gaussians with a separate convolutional deformation branch conditioned on FLAME expression and jaw parameters. FlexAvatar~\citep{flexavatar} reconstructs a head from a single image and regenerates the attributes of all Gaussians at every frame with a transformer decoder conditioned on a FLAME-derived expression code. DynaAvatar~\citep{dynavatar} reconstructs a full-body avatar with clothing dynamics from a single image and regenerates the Gaussians at every frame with a dynamic transformer conditioned on the current and 15 preceding SMPL-X poses. See Appendix~\ref{sec:supp_hosts} for more details.

\paragraph{Implementation details.}
The coefficient network of every host model is a shallow MLP, and for DynaAvatar it also depends on the window of preceding SMPL-X poses that the host model reads. The evaluation protocol of FlexAvatar fits the identity representation $w$ to the input images of every test subject, and our model follows it with a per-subject \emph{enrollment} that fine-tunes the coefficient network on outputs of the host model. We measure speed as the CPU time of one animation step, \ie the computation of the Gaussian attributes $\hat{G}$ from the animation parameters $\theta$ without rendering, on the same desktop CPU for all methods. Appendix~\ref{sec:supp_impl} and Table~\ref{tab:instantiation} give the settings of every host model.

\paragraph{Training data.}
Our method generates training data by running host models on samples of their inputs $(w_i,\theta_i)$. We take these samples from a subset of the training data of the corresponding host models and use them both to construct the basis and to train the coefficient network. For FlexAvatar we use 1{,}200 training identities with 8 expressions each, and for DynaAvatar we use frames sampled from its training sequences. AGORA is a generative model, hence we sample 10{,}000 identities from its latent space and combine each of them with 6 expressions of its training set FFHQ~\citep{FFHQ}. The results in Tables~\ref{tab:main}--\ref{tab:abl_allocation} and Figure~\ref{fig:pareto} are reported for identities excluded from this data.

\subsection{Results}
\label{subsec:quantitative}

\begin{table}[t]
\centering
\caption{Comparison to the state of the art on the benchmark of each host model. In every panel the first row is the host model and the last row is the same model distilled by our method (+GALA). The rows in between are the methods compared in the paper of each host model, with the numbers reported there. The AGORA row is taken from \citet{agora}. The FlexAvatar and DynaAvatar rows are measured by us on the authors' released checkpoints, following their evaluation protocols (Appendix~\ref{sec:supp_eval}). \textbf{Best} and \underline{second-best} results, excluding the host model. EG3D and GGHead are not animatable. FPS is measured on a desktop GPU with a precomputed neutral avatar and mFPS on a mobile phone in a web browser. $^\ast$As reported in the model's paper.}
\label{tab:main}
\begin{minipage}[t]{0.567\linewidth}
\centering
{\scriptsize\textbf{(a) Face 3D GANs on FFHQ}\par}
\smallskip
\scriptsize
\setlength{\tabcolsep}{1.6pt}
\begin{tabular}{@{}lccccccc@{}}
\toprule
 & FID$\downarrow$ & AED$\downarrow$ & AED-jaw$\downarrow$ & ID$\uparrow$ & APD$\downarrow$ & FPS$\uparrow$ & mFPS$\uparrow$ \\
\midrule
AGORA   & 3.17 & 0.682 & 0.021 & 0.75 & 0.025 & 250$^\ast$ & 1 \\
\midrule
EG3D   & \underline{3.28} & -- & -- & -- & -- & -- & -- \\
GGHEAD & 4.06 & -- & -- & -- & -- & -- & -- \\
Next3D & \textbf{3.18} & 0.930 & 0.046 & \underline{0.74} & 0.031 & 15$^\ast$ & -- \\
GAIA   & 3.85 & \textbf{0.530} & \underline{0.040} & 0.72 & \underline{0.027} & \underline{52}$^\ast$ & -- \\
\midrule
AGORA + GALA (ours) & 3.47 & \underline{0.705} & \textbf{0.022} & \textbf{0.76} & \textbf{0.026} & \textbf{1{,}275} & \textbf{60} \\
\bottomrule
\end{tabular}
\end{minipage}\hfill
\begin{minipage}[t]{0.43\linewidth}
\centering
{\scriptsize\textbf{(b) Clothed full-body feed-forward on 4D-Dress}\par}
\smallskip
\scriptsize
\setlength{\tabcolsep}{1pt}
\begin{tabular}{@{}lccccc@{}}
\toprule
 & PSNR$\uparrow$ & SSIM$\uparrow$ & LPIPS$\downarrow$ & FPS$\uparrow$ & mFPS$\uparrow$ \\
\midrule
DynaAvatar & 23.86 & 0.959 & 0.063 & 0.3 & -- \\
\midrule
IDOL    & \underline{21.31} & 0.948 & \underline{0.077} & -- & -- \\
PERSONA & 19.46 & 0.943 & 0.098 & -- & -- \\
LHM     & 21.03 & \underline{0.950} & 0.085 & -- & -- \\
\midrule
\makecell[l]{DynaAvatar\\+ GALA (ours)} & \textbf{23.22} & \textbf{0.957} & \textbf{0.066} & \textbf{1{,}284} & \textbf{60} \\
\bottomrule
\end{tabular}
\end{minipage}

\par\medskip
{\scriptsize\textbf{(c) One-shot face feed-forward}\par}
\smallskip
\scriptsize
\setlength{\tabcolsep}{1.15pt}
\begin{tabular}{@{}l cc ccc ccc ccc ccc@{}}
\toprule
 & \multicolumn{2}{c}{} & \multicolumn{6}{c}{Portrait animation (VFHQ)} & \multicolumn{6}{c}{3D avatar creation (Ava256)} \\
\cmidrule(lr){4-9}\cmidrule(lr){10-15}
 & \multicolumn{2}{c}{Speed} & \multicolumn{3}{c}{Self reenactment} & \multicolumn{3}{c}{Cross reenactment} & \multicolumn{3}{c}{Single-shot} & \multicolumn{3}{c}{Few-shot} \\
\cmidrule(lr){2-3}\cmidrule(lr){4-6}\cmidrule(lr){7-9}\cmidrule(lr){10-12}\cmidrule(lr){13-15}
 & FPS$\uparrow$ & mFPS$\uparrow$ & PSNR$\uparrow$ & LPIPS$\downarrow$ & AED$\downarrow$ & CSIM$\uparrow$ & AED$\downarrow$ & APD$\downarrow$ & PSNR$\uparrow$ & AKD$\downarrow$ & CSIM$\uparrow$ & PSNR$\uparrow$ & AKD$\downarrow$ & CSIM$\uparrow$ \\
\midrule
FlexAvatar & 140.7 & -- & 23.29 & 0.099 & 0.075 & 0.652 & 0.226 & 0.025 & 19.28 & 7.48 & 0.508 & 21.50 & 6.17 & 0.755 \\
\midrule
Portrait4D-v2 & -- & -- & 21.34 & 0.144 & 0.117 & \underline{0.656} & 0.268 & 0.273 & 11.9 & 7.77 & \textbf{0.578} & -- & -- & -- \\
GAGAvatar & -- & -- & 21.83 & \underline{0.122} & 0.111 & 0.633 & 0.253 & \underline{0.247} & 12.7 & \textbf{7.45} & \underline{0.555} & -- & -- & -- \\
LAM & \underline{280}$^\ast$ & \underline{35}$^\ast$ & \textbf{22.65} & \textbf{0.109} & \underline{0.102} & 0.651 & \underline{0.250} & 0.356 & \underline{13.1} & 11.2 & 0.411 & -- & -- & -- \\
GPAvatar & -- & -- & 21.04 & 0.150 & 0.132 & 0.564 & 0.255 & 0.328 & -- & -- & -- & 20.0 & \underline{5.72} & 0.341 \\
Avat3r & -- & -- & -- & -- & -- & -- & -- & -- & -- & -- & -- & \underline{20.8} & \textbf{5.66} & \underline{0.616} \\
\midrule
FlexAvatar + GALA (ours) & \textbf{1{,}296} & \textbf{60} & \underline{22.31} & \textbf{0.109} & \textbf{0.091} & \textbf{0.680} & \textbf{0.245} & \textbf{0.029} & \textbf{19.25} & \underline{7.52} & 0.491 & \textbf{21.31} & 6.31 & \textbf{0.740} \\
\bottomrule
\end{tabular}
\end{table}

\paragraph{Comparison to the state of the art.}
We evaluate every distilled model on the benchmark of its host model~\citep{agora,dynavatar,flexavatar} and denote a host model distilled by our method with +GALA. We adopt metrics from previous publications and measure image quality (FID, PSNR, SSIM, LPIPS), the accuracy of expressions, poses and keypoints (AED, AED-jaw, APD, AKD) and the preservation of identity (ID, CSIM). Appendix~\ref{sec:supp_eval} defines the metrics and the evaluation protocols.

Table~\ref{tab:main} shows that all three distilled models closely follow their host models and outperform the competing methods on most metrics. For AGORA, FID increases from 3.17 to 3.47, while the animation and identity metrics remain within 0.023 of the host model and are the best (AED-jaw, ID, APD) or second best (AED) among the other animatable models. For DynaAvatar, our model has a 0.64~dB lower PSNR than its host and outperforms all competing methods on all three quality metrics, \eg by more than 1.9~dB in PSNR. For FlexAvatar, our model obtains the best results on 8 of the 12 quality metrics, including one tie. Only LAM is competitive on portrait animation, where it has a 0.34~dB higher PSNR and the same LPIPS in self-reenactment. On Ava256 single-shot, our model stays within 0.03~dB PSNR of its host, which outperforms the compared one-shot methods by more than 3.8~dB PSNR under its published protocol~\citep{flexavatar}. The distilled model therefore retains this lead. With respect to its host, our model stays within 1~dB PSNR in self-reenactment and within 0.03~dB (single-shot) and 0.2~dB (few-shot) on Ava256.

At the same time, for the three tested host models our method reduces CPU time of their animation steps as 154$\rightarrow$5.5ms, 234$\rightarrow$4.4ms and 42.9s$\rightarrow$6.1ms (16.1ms with skinning included) %
respectively (Table~\ref{tab:instantiation}). This corresponds to speed-ups of about $\times28$, $\times54$ and $\times7{,}000$ times, or $\times2{,}700$ times for DynaAvatar when its skinning is included. With an animation step below 7ms on a desktop CPU for all three host models, our method enables real-time animation, even on a mobile phone, for host models that were not capable of it before. AGORA runs at 1 frame per second on the phone, and DynaAvatar, which models garment dynamics, reaches only 0.3 frames per second even on a desktop GPU, whereas our distilled models run at 60 frames per second on the phone (mFPS in Table~\ref{tab:main}). The \textbf{video on our \href{https://ramazan793.github.io/gala/}{project page}} demonstrates all three distilled models running in the web browser of a mobile phone (Appendix~\ref{sec:supp_demo}).

\paragraph{Qualitative results.}
\label{subsec:qualitative}

\begin{figure}[t]
    \centering
    \includegraphics[width=\linewidth]{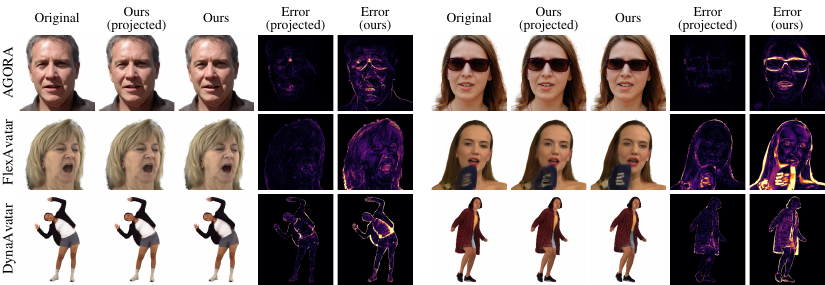}
    \caption{Qualitative comparison with the host models (\emph{Original}), two examples per host model. \emph{Ours (projected)} uses projected coefficients $c^\star$, which require running the host model and show what the basis can express (Section~\ref{subsubsec:regression}), and \emph{Ours} the coefficients predicted by the coefficient network. Error maps show the absolute difference to the host model per pixel, averaged over the RGB channels (values in $[0,1]$), from 0 (black) to 0.2 (yellow). Zoom in for details.}
    \label{fig:qualitative}
\end{figure}

Figure~\ref{fig:qualitative} compares renderings of our method and of the host models for two examples per host model with pronounced facial expressions and body motion, and shows where the errors of our method are located. Our method faithfully reproduces both the facial expressions and the garment motion. With projected coefficients, it reproduces the host models with hardly visible errors in these examples. With predicted coefficients, which add the error of coefficient prediction, the remaining errors are small and localized at thin structures and at the outlines of the silhouettes, such as teeth and glasses for AGORA, mouth interior and hair boundary for FlexAvatar, and garment boundaries for DynaAvatar, and are mainly revealed by the error maps. Appendix~\ref{sec:supp_blendshapes} visualizes the local motions captured by individual blendshapes, and Appendix~\ref{sec:supp_qualitative} presents more difficult examples, including a subject for which the coefficient network fails.

\subsection{Ablation Study}
\label{subsec:ablations}

\begin{table}[t]
\centering
\caption{Ablation of the basis construction. Row 1 of (a) is simple PCA (Section~\ref{subsubsec:simplepca}).}
\label{tab:abl_ingredients}
\label{tab:abl_blocks}
\footnotesize
\setlength{\tabcolsep}{4.5pt}
\begin{minipage}[t]{0.6\linewidth}
\centering
{\textbf{(a) Design choices of our method}\par}\smallskip
\begin{tabular}{@{}ccc ccc@{}}
\toprule
Per-attribute & \multicolumn{2}{c}{Rendering-aware} & \multicolumn{3}{c}{PSNR to host $\uparrow$} \\
\cmidrule(lr){2-3}\cmidrule(lr){4-6}
bases & PCA & allocation & AGORA & Flex & Dyna \\
\midrule
$\times$ & $\times$ & $\times$ & 32.77 & 21.56 & 21.38 \\
\midrule
$\times$ & \checkmark & \checkmark & 37.83 & 29.48 & 25.86 \\
\checkmark & $\times$ & \checkmark & 42.32 & 32.74 & 34.04 \\
\checkmark & \checkmark & $\times$ & 37.23 & 30.39 & 21.34 \\
\midrule
\checkmark & \checkmark & \checkmark & \textbf{42.57} & \textbf{33.09} & \textbf{34.62} \\
\bottomrule
\end{tabular}
\end{minipage}\hfill
\begin{minipage}[t]{0.38\linewidth}
\centering
{\textbf{(b) Number of blocks}\par}\smallskip
\begin{tabular}{@{}lccc@{}}
\toprule
 & \multicolumn{3}{c}{PSNR to host $\uparrow$} \\
\cmidrule(lr){2-4}
$B$ & AGORA & Flex & Dyna \\
\midrule
1  & 34.35 & 26.19 & 26.04 \\
4  & 37.00 & 28.80 & 28.63 \\
8  & 38.04 & 30.13 & 30.64 \\
16 & 40.46 & 31.54 & 32.65 \\
32 (ours) & \textbf{42.57} & \textbf{33.09} & \textbf{34.62} \\
\bottomrule
\end{tabular}
\end{minipage}
\end{table}

\paragraph{Design choices of the basis construction.}
We evaluate the representation power of our basis, \ie how well it reconstructs the output of the host model independently of coefficient prediction, by the PSNR between the renderings of our model with projected coefficients $c^\star$ and of the host model on held-out identities, at the same memory budget for all compared variants. Table~\ref{tab:abl_ingredients}a compares our full method with simple PCA, the single global basis of Section~\ref{subsubsec:simplepca} with the rank given by the budget, and with variants that revert one design choice while the others are kept. Simple PCA is 9.7, 11.5 and 13.2~dB below our full method for AGORA, FlexAvatar and DynaAvatar. Most of this gap comes from the two design choices that decide how the memory is spent: a joint basis over all attributes of a block costs 4.7, 3.6 and 8.8~dB, since a joint component stores all $d$ channels instead of $d_a$ and couples attributes with different scales, and scoring the components by explained variance instead of the value $v_{a,j,k}$ costs 5.3, 2.7 and 13.3~dB, since explained variance is not comparable between attribute groups with different units and spends most of the memory on attributes with little effect on the image. Rendering-aware PCA itself improves the directions of the basis by a further 0.25, 0.35 and 0.58~dB over Euclidean PCA ($\beta=\gamma=0$) with the same numbers of components.

\begin{wraptable}{r}{0.53\linewidth}
\centering
\footnotesize
\setlength{\tabcolsep}{3pt}
\caption{Allocation of components, PSNR to the host model.}
\label{tab:abl_allocation}
\begin{tabular}{@{}lccc@{}}
\toprule
Rank & AGORA & Flex & Dyna \\
\midrule
Global, $K_{a,j}=K$ & 36.39 & 31.91 & 26.17 \\
Per group, $K_{a,j}=K_a$ & 39.19 & 32.97 & 33.95 \\
Per group and block, $K_{a,j}$ (ours) & \textbf{42.57} & \textbf{33.09} & \textbf{34.62} \\
\bottomrule
\end{tabular}
\end{wraptable}
\paragraph{Allocation of components.}
Table~\ref{tab:abl_allocation} evaluates the need for allocating the numbers of components $K_{a,j}$ individually. We compare our greedy allocation (Section~\ref{subsubsec:improvedpca}) with two simpler rules for the same bases and memory budget: one rank $K$ shared by all attribute groups and blocks, and one rank $K_a$ per attribute group shared by all blocks. Table~\ref{tab:abl_ingredients}a changes the values of the components, whereas here the values are fixed and only the constraints on the ranks change. A single rank reduces PSNR by 6.2, 1.2 and 8.5~dB, since it assigns the same number of components to every attribute group and to every block irrespective of their motion. Individual ranks per attribute group recover a large part of this gap. Our allocation per attribute group and block improves PSNR by a further 3.4, 0.1 and 0.7~dB. Given the values of the components, the greedy allocation requires no tuning and no rendering, and takes a few seconds.

\paragraph{Number of spatial blocks.}
Table~\ref{tab:abl_ingredients}b varies the number of spatial blocks $B$ for a fixed basis size. Compared with a global basis per attribute group ($B=1$), $B=32$ local blocks improve PSNR by 8.2, 6.9 and 8.6~dB. A component of a local basis stores $n_jd_a$ instead of $nd_a$ values, hence more components fit into the same memory and they can be concentrated in the regions with complex motion. PSNR increases monotonically with $B$ over the tested range, and we use $B=32$ in all other experiments. For larger $B$ we observed no consistent improvement with predicted coefficients, since more components make the coefficients harder to regress.

\begin{figure}[t]
\centering
\includegraphics[width=\linewidth]{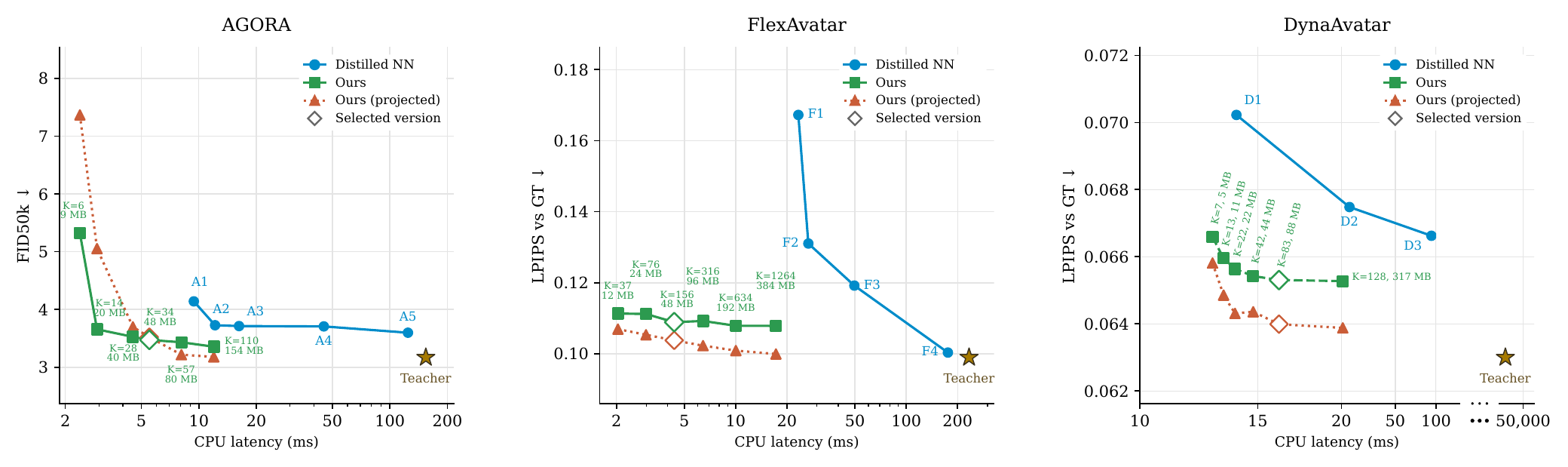}
\caption{Comparison to alternative distillation: quality against the CPU time of the animation step, lower is better on both axes. \emph{Teacher} is the host model, our models are labeled with the components per Gaussian and the basis size, and the diamond marks our final model.}
\label{fig:pareto}
\end{figure}

\paragraph{Comparison to alternative distillation.}
To further demonstrate the value of our linear blendshape distillation approach, we compare it to a generic method for neural network distillation. Knowledge distillation~\citep{bucilua2006compression,hinton2015distilling} trains a smaller student network to reproduce the outputs of a frozen teacher on the inputs of the teacher. We distill students that are thinner and shallower copies of the animation network of corresponding host models. We train students of multiple sizes using the same training data as we use for training our method.
Figure~\ref{fig:pareto} compares this standard distillation with GALA for a range of memory budgets. For all three host models, none of the distilled networks approaches our method in terms of speed and accuracy.
Additional details of this study are available in Appendix~\ref{sec:supp_ugb}.

%% file: sec/6_conclusion.tex
\section{Limitations and Future Work}
\label{sec:limitations}

While GALA already shows excellent results, its coefficient prediction leaves room for improvement.  
With projected coefficients $c^\star$ our models reproduce their host models closely, whereas with predicted coefficients their PSNR with respect to the host model is lower by 5.7~dB for AGORA and by 12.3~dB for FlexAvatar at a memory budget of 48~MB, and a larger budget mainly improves the results with projected coefficients (Figure~\ref{fig:pareto}). 
Another direction for improvement concerns the spatial partition, which is currently fixed to $B=32$ blocks for all host models. Meanwhile an optimal partition may depend on the motion and could be selected jointly with the numbers of components $K_{a,j}$. Finally, our method inherits the limitations of host models, and our results suggest that avatar models could be trained directly with an \mbox{identity-shared blendshape basis, which we leave for future work.}

\section{Conclusion}
\label{sec:conclusion}

We introduced GALA, which replaces per-frame neural decoding of pretrained Gaussian avatar models with a shallow coefficient network and a linear combination of blendshapes. To the best of our knowledge, GALA is the first method to extract and leverage a blendshape basis shared across identities%
, whereas previous Gaussian blendshape methods require a separate basis for every subject. 
We apply GALA to three diverse avatar models for head and full-body animation and demonstrate excellent computational gains while closely preserving the quality of original models and outperforming recent baselines. GALA requires no retraining of original models, it accelerates CPU execution up to three orders of magnitude and enables animation to run at 60~fps on a mobile phone. These results demonstrate the effectiveness and broad applicability of proposed Gaussian blendshape distillation for efficient, high-fidelity on-device avatar animation.

\subsection*{Acknowledgements}
This work was in-part supported by the Institute of Information \& communications Technology Planning \& Evaluation (IITP) grant funded by the Korean government(MSIT) (No. RS-2024-00457882, National AI Research Lab Project).

%% file: sec/X_suppl.tex
Appendix~\ref{sec:supp_metric} details the estimation of the rendering-aware metric and the construction of the basis in this metric space. Appendix~\ref{sec:supp_hosts} describes the three host models and Appendix~\ref{sec:supp_demo} our mobile demos. Appendix~\ref{sec:supp_impl} gives the implementation details of our method, Appendix~\ref{sec:supp_coeffnet} describes the coefficient networks, and Appendix~\ref{sec:supp_eval} provides the evaluation details, including the evaluation of FlexAvatar on Ava256. Appendix~\ref{sec:supp_ugb} compares our method with distilled networks. Appendix~\ref{sec:supp_per_subject} compares the distilled FlexAvatar and DynaAvatar models with per-subject blendshape avatars. Appendix~\ref{sec:supp_qualitative} adds qualitative results, and Appendix~\ref{sec:supp_blendshapes} visualizes the motions of individual blendshapes. Please refer to the video on our \href{https://ramazan793.github.io/gala/}{project page} for more.

\section{Rendering-Aware Metric and Basis Construction}
\label{sec:supp_metric}

\paragraph{Estimation of the rendering Hessian.}
The rendering-aware metric of Section~\ref{subsubsec:improvedpca} is based on the Gauss--Newton Hessian $H=J^\top J$ of the rendered image with respect to the Gaussian attributes, which is too large to construct. We approximate it by its per-Gaussian diagonal blocks within each attribute group, which preserve the interactions among the channels of one Gaussian and one attribute group and discard the interactions between Gaussians and between attribute groups. We estimate these blocks with a stochastic estimator in the spirit of the trace and diagonal estimators of \citet{hutchinson1989stochastic} and \citet{bekas2007estimator}. For a random vector $\epsilon$ in image space with entries $\pm1$ of equal probability, one backward pass through the renderer gives
\begin{equation}
    g=J^\top\epsilon,
    \qquad
    \mathbb{E}[gg^\top]=J^\top\mathbb{E}[\epsilon\epsilon^\top]J=J^\top J.
\end{equation}
For every Gaussian and attribute group $a$ we therefore accumulate the $d_a\times d_a$ block of $gg^\top$ that belongs to the $d_a$ channels of this Gaussian, and average over identities, animation parameters and cameras of the training samples, with 16 probes per frame. The estimate is symmetrized and normalized to unit mean trace per channel within each attribute group, which gives $\widehat H_a$. The normalization constant $\bar h_a$, the mean trace per channel before normalization, is used again by the allocation of components.
The landmark matrix $\widehat H_a^{\mathrm{lmk}}$ is estimated in the same way, with the Jacobian of the landmark positions in place of $J$.

\paragraph{PCA in the metric space.}
For attribute group $a$ and block $j$, let $M_{a,j}$ be the restriction of $M_a$ of Eq.~\ref{eq:metric} to the $n_j$ Gaussians of the block, and let $R_{a,j}\in\mathbb{R}^{n_jd_a\times N}$ contain the corresponding residuals as columns. The local basis is the solution of
\begin{equation}
    \min_{U_{a,j},\{c_{i,a,j}\}}
    \sum_{i=1}^{N}\|r_{i,a,j}-U_{a,j}c_{i,a,j}\|_{M_{a,j}}^2,
    \qquad
    \|z\|_M^2=z^\top Mz,
    \qquad
    U_{a,j}^\top M_{a,j}U_{a,j}=I.
\label{eq:metricpca}
\end{equation}
Substituting $y=M_{a,j}^{1/2}r$ turns the objective into the Euclidean objective of Eq.~\ref{eq:simplepcabasis}, hence the problem is solved by uncentered PCA of
\begin{equation}
    Y_{a,j}=M_{a,j}^{1/2}R_{a,j}.
\end{equation}
If $V_{a,j}$ contains the top $K_{a,j}$ left singular vectors of $Y_{a,j}$, the basis and the projected coefficients are
\begin{equation}
    U_{a,j}=M_{a,j}^{-1/2}V_{a,j},
    \qquad
    c_{a,j}^\star=U_{a,j}^\top M_{a,j}r_{a,j},
\label{eq:metricbasis}
\end{equation}
so that $U_{a,j}^\top M_{a,j}U_{a,j}=I$. Since $M_{a,j}$ is block diagonal with one $d_a\times d_a$ block per Gaussian, $M_{a,j}^{1/2}$ and $M_{a,j}^{-1/2}$ are computed per Gaussian. Setting $\beta=\gamma=0$ gives $M_{a,j}=\alpha I$ and reduces the construction to the simple PCA of Section~\ref{subsubsec:simplepca} applied to the block.

\paragraph{Value of a component.}
The allocation of Section~\ref{subsubsec:improvedpca} scores component $k$ of attribute group $a$ and block $j$ by
\begin{equation}
    v_{a,j,k}=\frac{\bar h_a}{\beta}\lambda_{a,j,k},
\label{eq:componentvalue}
\end{equation}
for $\beta>0$, where $\lambda_{a,j,k}$ is the $k$th squared singular value of $Y_{a,j}$ and $\bar h_a$ is the mean trace per channel of $\widehat H_a$ before normalization. Since $\widehat H_a$ is normalized separately for every attribute group, the singular values of different groups are not directly comparable, and the factor $\bar h_a/\beta$ undoes the normalization and the weight $\beta$ of the rendering term. The value then combines the estimated reduction of the squared image error with the weighted reductions of the squared attribute error and, if $\gamma>0$, of the landmark error, summed over the samples. Since the values are non-increasing within each local basis, the greedy allocation by value per byte retains the components of every local basis in spectral order.

\section{Host Models}
\label{sec:supp_hosts}

We apply our method to three host models, spanning different architectures, training objectives, and animation mechanisms. Together, these state-of-the-art models cover dedicated deformation networks, decoders with integrated expression conditioning, and temporally conditioned full-body animation. Table~\ref{tab:instantiation} lists the numbers of Gaussians and the attribute groups of the host models together with the settings of our distilled models.

\smallskip\noindent
\textbf{AGORA~\citep{agora}} is a Gaussian avatar GAN trained adversarially on unstructured images of human heads. It follows the dual-branch design of Next3D~\citep{Next3D} and GAIA~\citep{gaia}, using GGHead's UV-space Gaussian representation~\citep{gghead}. A mapping network produces an identity code $w$, and an identity branch generates canonical Gaussian attributes on a UV grid aligned with FLAME. A separate deformation branch, conditioned on $w$ and FLAME expression and jaw parameters $\theta$, predicts attribute residuals. The identity branch is evaluated once per subject, whereas the deformation branch is evaluated at every frame. Identities can be generated by latent sampling or obtained by image inversion.

\smallskip\noindent
\textbf{FlexAvatar~\citep{flexavatar}} reconstructs an animatable head from a single image using an encoder--decoder trained with rendering losses on monocular and multi-view portrait videos. The encoder produces an identity representation in template UV space. A FLAME-derived expression code conditions the decoder through cross-attention. Each animation frame therefore requires evaluating the full decoder to regenerate the attributes of all Gaussians. This model tests whether distillation remains effective when animation is integrated into the decoder rather than implemented by a separate deformation branch.

\smallskip\noindent
\textbf{DynaAvatar~\citep{dynavatar}} reconstructs an animatable full-body avatar with clothing dynamics from a single image. Image encoders and a static transformer produce identity features on the SMPL-X template once per subject. A dynamic transformer, conditioned on a window of the current and 15 preceding SMPL-X poses, regenerates the attributes of template-anchored Gaussians at every frame, which are then posed by linear blend skinning. This model tests our method on full-body motion with garment dynamics that depend on the pose history. Its neural animation is by far the most expensive of the three models (Table~\ref{tab:instantiation}).

\section{Mobile Browser Demo}
\label{sec:supp_demo}

\begin{figure}[htb]
\centering
\begin{minipage}[c]{0.33\linewidth}
\centering
\includegraphics[width=\linewidth]{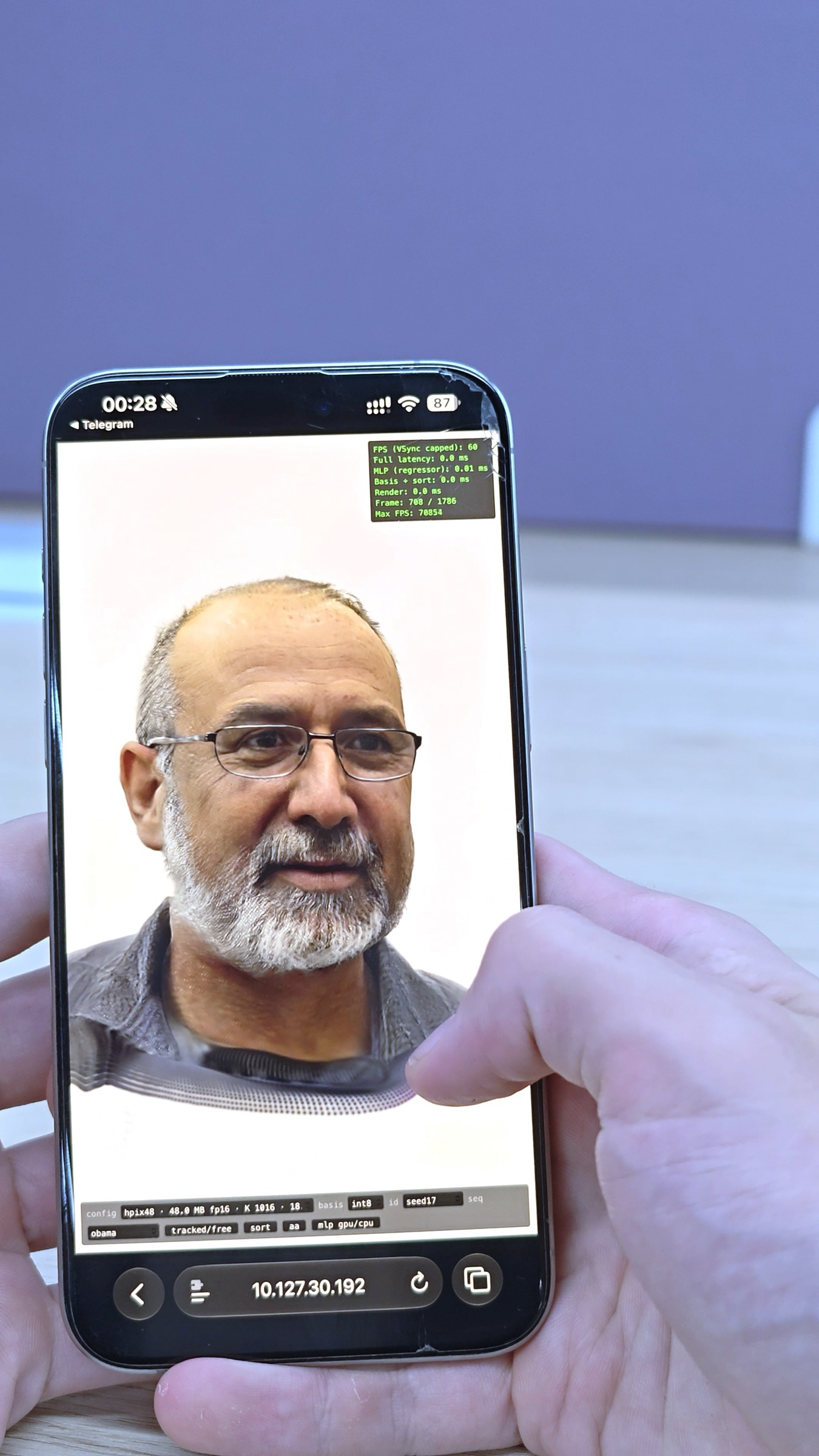}
\end{minipage}\hspace{0.012\linewidth}%
\begin{minipage}[c]{0.52\linewidth}
\centering
\includegraphics[width=\linewidth]{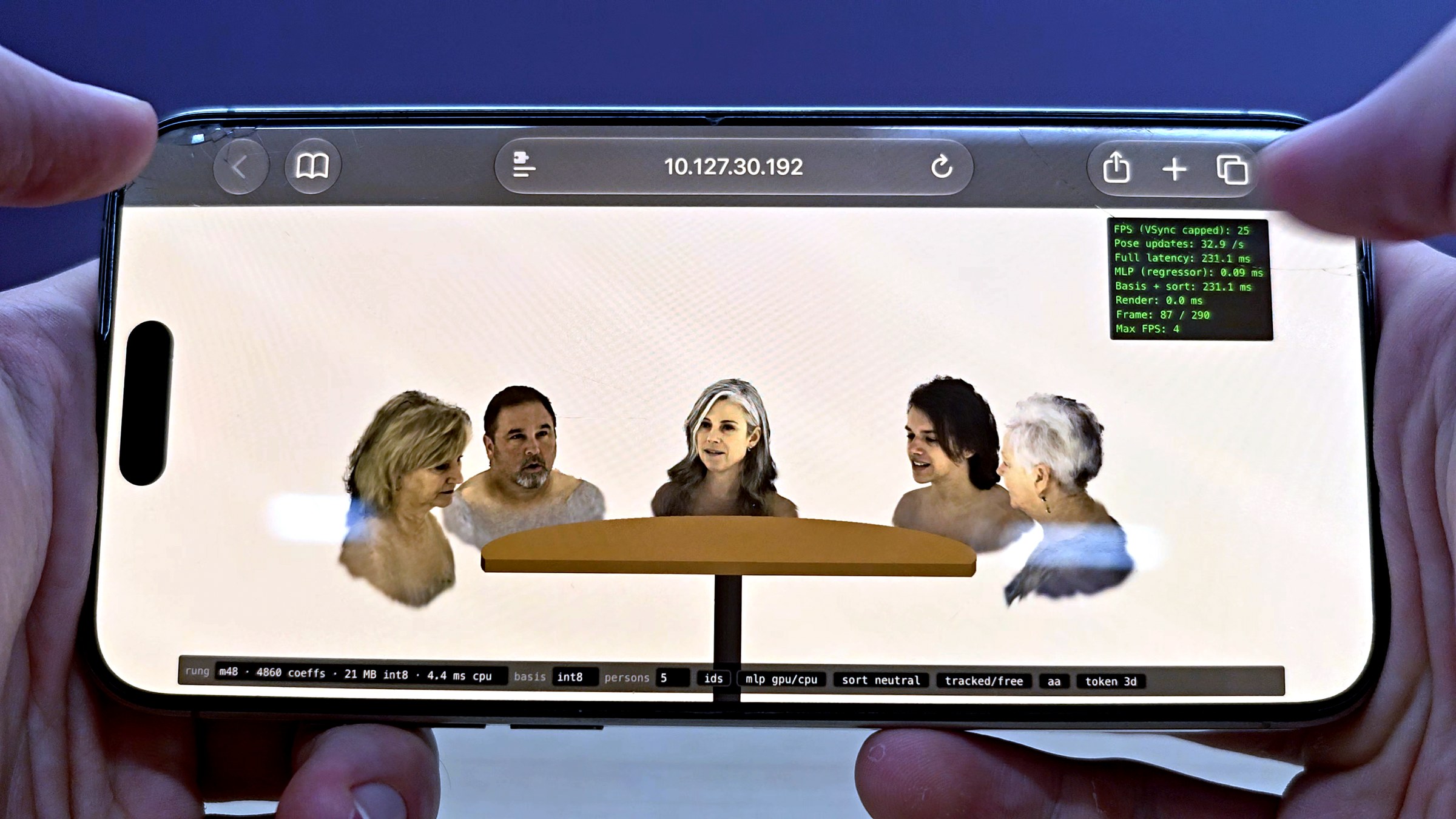}\\[3pt]
\includegraphics[width=\linewidth]{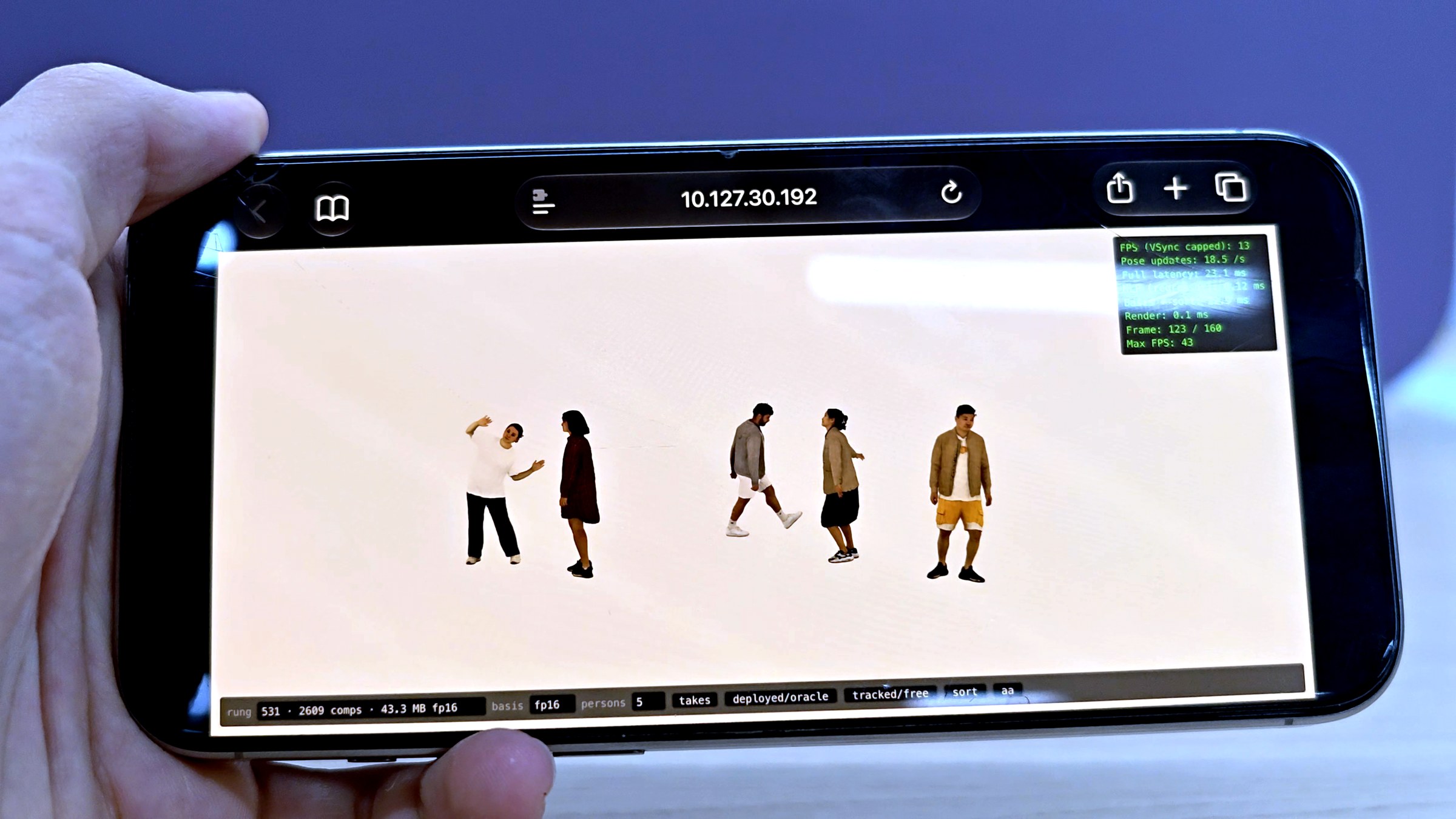}
\end{minipage}
\caption{Our browser demos on a mobile phone, screenshots from the \textbf{video on our \href{https://ramazan793.github.io/gala/}{project page}}. Left: AGORA + GALA driven by the tracked expressions of a video. Right: five FlexAvatar + GALA avatars, each reconstructed from one photo, around a table (top), and five DynaAvatar + GALA full-body avatars with clothing dynamics (bottom). The overlays show the frame rate and the timing of the animation step.}
\label{fig:demo}
\end{figure}

Our real-time demos run in a web browser on desktop and mobile devices (Figure~\ref{fig:demo}). They are built on the open-source WebGL Gaussian splat viewer of \citet{kwok2023splat}, whose depth sort and splat rasterizer we keep and whose static point cloud we replace by our animated avatar. The neutral avatar and the blendshape basis are loaded once as textures. At every frame, the browser reads the animation parameters of the driving sequence, evaluates the coefficient network as fragment-shader passes on the GPU, forms the Gaussian attributes as the neutral avatar plus the coefficient-weighted sum of the blendshapes in a second pass, and hands the result to the sort and the rasterizer of the viewer. No host network runs in the browser. The FlexAvatar and DynaAvatar demos can show several avatars at once. The AGORA and FlexAvatar demos use the same viewer and differ only in the animation parameters they read, the FLAME expression and jaw parameters and the tracked expression code respectively, and in the sizes of the basis and the network (Table~\ref{tab:instantiation}). The \textbf{video on our \href{https://ramazan793.github.io/gala/}{project page}} shows the demos in action.

\section{Implementation Details}
\label{sec:supp_impl}

\begin{table}[htb]
\centering
\caption{Host models and the settings of our distilled models. Components are listed in total and per attribute group, in the order of the attribute groups. The animation step is timed on a CPU without rendering. For DynaAvatar, both the host model and ours additionally pose the Gaussians by linear blend skinning, which takes 10~ms per frame.}
\label{tab:instantiation}
\footnotesize
\setlength{\tabcolsep}{4pt}
\renewcommand{\arraystretch}{1.05}
\begin{tabular}{@{}l >{\raggedright\arraybackslash}p{0.16\linewidth} >{\raggedright\arraybackslash}p{0.28\linewidth} >{\raggedright\arraybackslash}p{0.28\linewidth}@{}}
\toprule
 & AGORA & FlexAvatar & DynaAvatar \\
\midrule
Gaussians $n$ & 228{,}693 & 58{,}361 & 40{,}000 \\
attribute groups $a$ & position, scale, rotation & position, scale, rotation, opacity, color & position, scale, rotation, opacity, color, skinning weights \\
animation parameters $\theta$ & FLAME expression and jaw & FLAME-derived expression code & current and 15 preceding SMPL-X poses \\
\midrule
spatial blocks $B$ & 32 & 32 & 32 \\
metric $\alpha$, $\beta$, $\gamma$ & 1, 16, 0 & 1, 16, 24 & 1, 64, 0 \\
bytes per value $b$ & 2 & 2 & 4 \\
components $K$ & 1{,}016 \newline (1{,}005 / 4 / 7) & 4{,}860 \newline (3{,}074 / 197 / 174 / 884 / 531) & 2{,}686 \newline (2{,}116 / 76 / 7 / 1 / 302 / 184) \\
memory budget $\mathcal{B}$ & 48.0 MB & 48.0 MB & 88.3 MB \\
coefficient network & 1.6 MB & 5.9 MB & 60.7 MB \\
\midrule
animation step, host & 154 ms & 234 ms & 42{,}917 ms \\
animation step, ours & 5.5 ms & 4.4 ms & 6.1 ms \\
\bottomrule
\end{tabular}
\end{table}

\paragraph{Basis construction.}
We use the same configuration of our method for all host models unless stated otherwise. The Gaussians are partitioned into $B=32$ spatial blocks, and a separate basis is constructed for every attribute group and block (Section~\ref{subsubsec:improvedpca}). The rendering-aware metric $M_a$ (Eq.~\ref{eq:metric}) uses $\alpha=1$ and $\beta=16$, with $\beta=64$ for DynaAvatar, and is estimated with 16 random probes per frame (Appendix~\ref{sec:supp_metric}). The landmark term is used only for the positions of FlexAvatar ($\gamma=24$), which is itself trained with a landmark loss. The numbers of components $K_{a,j}$ are allocated under a memory budget $\mathcal{B}$ on the storage of the basis (Section~\ref{subsubsec:improvedpca}), which is listed in Table~\ref{tab:instantiation}. The basis is stored in half precision ($b=2$) for the two head models and in single precision ($b=4$) for DynaAvatar.

\paragraph{Coefficient network.}
The architecture and training of the coefficient network $f_\phi$ (Section~\ref{subsubsec:regression}) are given in Appendix~\ref{sec:supp_coeffnet}. Table~\ref{tab:instantiation} lists its size for every host model.

\paragraph{Enrollment for FlexAvatar.}
The evaluation protocol of FlexAvatar fits the identity representation $w$ to the input images of every test subject. Our model follows this protocol with a per-subject enrollment: given the fitted $w$ and the expressions of the input images, we estimate a constant correction of the neutral avatar and fine-tune the coefficient network for 200 steps on outputs of the host model. The enrollment uses no target frames and no rendering.

\paragraph{Timing.}
We measure speed as the CPU time of one animation step, \ie the computation of the Gaussian attributes $\hat{G}$ from the animation parameters $\theta$ without rendering. All methods are timed on the same CPU, an Intel Core Ultra 9 285K, with eight threads and batch size one. For DynaAvatar, the linear blend skinning of 10~ms per frame is excluded from the animation step in Table~\ref{tab:instantiation} and included for all methods in Figure~\ref{fig:pareto}, whose time axis is cut before the host model at 43~s.

\section{Coefficient Networks}
\label{sec:supp_coeffnet}

Table~\ref{tab:coeffnet} summarizes the coefficient networks $f_\phi$ (Section~\ref{subsubsec:regression}) of the three distilled models of Table~\ref{tab:main}.

\begin{table}[htb]
\centering
\caption{Coefficient networks of the distilled models of Table~\ref{tab:main}. Hidden layers are listed with their widths. Parameters and multiply-accumulate operations (MAC) are those of one animation step of the network, without the basis. The FlexAvatar identity encoder (6.4~M parameters) runs once per identity and is not included.}
\label{tab:coeffnet}
\footnotesize
\setlength{\tabcolsep}{4pt}
\renewcommand{\arraystretch}{1.05}
\begin{tabular}{@{}l >{\raggedright\arraybackslash}p{0.18\linewidth} >{\raggedright\arraybackslash}p{0.22\linewidth} >{\raggedright\arraybackslash}p{0.30\linewidth}@{}}
\toprule
 & AGORA & FlexAvatar & DynaAvatar \\
\midrule
identity descriptor $w$ & latent of the neutral render (512) & embedding of the fitted avatar code (256) & identity tokens and shape parameters of the host encoder (3{,}082) \\
animation parameters $\theta$ & FLAME expression, jaw and eyelids (55) & articulation code (135) & 16 SMPL-X frames $\times$ 79 values \\
hidden layers & 256, 512 & 512, 512 & per frame 512, 512, two temporal convolutions of 512, then 1{,}024, 1{,}024 \\
activation & leaky ReLU & GELU & SiLU \\
outputs $K$ & 1{,}016 & 4{,}860 & 2{,}686 \\
parameters & 0.80~M & 2.96~M & 15.2~M \\
MAC per frame & 0.80~M & 2.96~M & 42~M \\
\midrule
training pairs (identities) & 57{,}000 (9{,}500) & 30{,}576 (2{,}184) & 20{,}700 (1{,}035) \\
held out for selection & 500 identities & 98 identities & 22 subjects \\
optimizer, learning rate & Adam, $10^{-3}$ & AdamW, $2\cdot 10^{-3}$ & AdamW, $10^{-3}$ \\
weight decay & 0 & $10^{-6}$ & $10^{-4}$ \\
batch size & 256 & all pairs & 256 \\
epochs run (selected) & 1{,}008 (8) & 512 (412) & 60 (40) \\
gradient norm clipping & 10 & 10 & none \\
training time, one GPU & 11\,min & 5\,min & 45\,s \\
per-subject step & none & enrollment & none \\
\bottomrule
\end{tabular}
\end{table}

\paragraph{Inputs and architecture.}
The identity descriptor $w$ is computed once per identity. For AGORA it is the latent code of the neutral render. For FlexAvatar it is an embedding of the avatar code that the host model fits to the input image, produced by a small encoder with layers of 512 and 256 units that is trained jointly with the network. For DynaAvatar it consists of the pooled identity tokens and the shape parameters of the host's identity encoder. The animation parameters $\theta$ are the ones the host model reads: the FLAME expression, jaw and eyelid parameters for AGORA, the articulation code for FlexAvatar, and the window of the current and 15 preceding SMPL-X frames for DynaAvatar. The two head models use an MLP with two hidden layers on the concatenation of $w$ and $\theta$. The DynaAvatar network embeds every frame of the window with two layers, applies two temporal convolutions with kernel size 3 over the window, concatenates the flattened result with an embedding of $w$, and predicts the coefficients with two hidden layers. Its output is anchored at a reference pose, $\widehat c=f_\phi(w,\theta)-f_\phi(w,\theta_{\mathrm{ref}})$, so that the reference pose gives the neutral avatar for every subject. The last layer of every network outputs the $K$ coefficients of all attribute groups and blocks.

\paragraph{Training.}
The targets are the projected coefficients $c^\star$ of the residuals of the host model with respect to the neutral avatar. The distillation loss of Section~\ref{subsubsec:regression} is normalized per attribute group by the coefficient energy of that group on the training data. Every training pair combines an identity with animation parameters taken from elsewhere: an FFHQ expression for AGORA, an articulation code sampled from all training identities for FlexAvatar, and a motion window of another training subject for DynaAvatar, whose 1{,}035 identity instances are single source images of the 26 4D-Dress subjects and 136 DNA-Rendering~\citep{dnarendering} actors in the training data of the host model. All networks are trained in single precision with a cosine learning-rate schedule, and the checkpoint is selected on held-out identities, never on frames of the training identities.

\section{Evaluation Details}
\label{sec:supp_eval}

\paragraph{Benchmarks and metrics.}
Every distilled model is evaluated on the benchmark of its host model. AGORA is evaluated on FFHQ~\citep{FFHQ} with the protocol of \citet{Next3D}. FID compares the distributions of generated and real images and measures image quality, AED and AED-jaw measure the accuracy of the expression and jaw animation, ID measures the consistency of the identity under animation, and APD measures the accuracy of the head pose. DynaAvatar is evaluated on 4D-Dress~\citep{dress4d}, where PSNR, SSIM and LPIPS measure the pixel-wise, structural and perceptual similarity of the animated avatar to ground-truth images. FlexAvatar is evaluated on portrait animation on VFHQ~\citep{VFHQ} and on 3D avatar creation from one or a few images on Ava256~\citep{ava256}. Here PSNR and LPIPS measure image quality with respect to the ground truth, AED, APD and the keypoint distance AKD measure animation accuracy, and CSIM, the cosine similarity of face recognition features, measures identity preservation. For AGORA, the rows of Table~\ref{tab:main}a are taken from \citet{agora}. For FlexAvatar and DynaAvatar, the host rows of Table~\ref{tab:main}b and c are measured by us on the checkpoints released by the authors, and the rows of the compared methods are quoted from the host papers.

\paragraph{Evaluation of DynaAvatar on 4D-Dress.}
We use the checkpoint released by \citet{dynavatar} and follow their evaluation protocol on 83 sequences and 12{,}200 frames of the 4D-Dress test split. Our measurement of the host, 23.86~dB PSNR, 0.959 SSIM and 0.063 LPIPS, reproduces the published numbers of 23.74~dB, 0.960 and 0.064 within 0.12~dB. The host and our model are evaluated on the same renders.

\paragraph{Evaluation of FlexAvatar.}
\label{sec:supp_ava256}
On VFHQ we use the authors' main model FLEX-1 (\texttt{ckpt-900k}) and follow their released VFHQ-Test protocol. On Ava256 we use the released checkpoint \texttt{FLEX-2-avat3r-eval}, which according to the repository was not trained on the Avat3r test identities. The host row, our basis and our coefficient network are all computed from this checkpoint. \citet{flexavatar} specify the structure of the Ava256 protocol, namely one frontal input frame with 10 expressions and 4 cameras per subject for the single-shot setting and four input images per subject following Avat3r for the few-shot setting, but not the particular subjects, sequences or expression frames, and the authors did not respond to our request for them. We therefore instantiate the protocol as follows. \citet{flexavatar} evaluate on 5 subjects that are not identified, so we use all 8 unseen subjects of the authors' released Avat3r evaluation specification and the \texttt{EXP\_free\_face} sequence of each subject. The input is the subject's most frontal calibrated camera at the first frame of the sequence. The targets are evenly spaced frames of the sequence excluding the input frame, each rendered from the subject's 4 Avat3r input cameras. The host and our model are evaluated on exactly the same identities, input images and target frames. Our numbers for the host consequently differ from those in \citet{flexavatar}, \eg 19.27 versus 16.9~dB single-shot PSNR over 8 and 5 subjects respectively, and the numbers of the compared methods in Table~\ref{tab:main}c are quoted from that paper. Since our method uses the host only through its outputs, the difference between the host and our model is expected to be unchanged under the original evaluation subset. The few-shot setting uses the authors' released evaluation specification for the Avat3r test identities~\citep{avat3r} with 8 subjects, 4 input views and 631 target images.

\paragraph{Ablations.}
Table~\ref{tab:supp_simplepca} compares the simple PCA basis with our full method after coefficient prediction, on the benchmarks of Table~\ref{tab:main}. A coefficient network of the same architecture is trained for each basis. Our method is better on every quality metric except the identity metrics, by 2.4~dB PSNR for AGORA, 0.9~dB for FlexAvatar and 0.6~dB for DynaAvatar, and the higher ID and CSIM of simple PCA are a sign of under-animation. For the second rule of Table~\ref{tab:abl_allocation}, the ranks $K_a$ per attribute group are chosen without rendering: among the splits of the memory budget between the attribute groups, we take the one that maximizes the total value $v_{a,j,k}$ of the retained components, the same rendering-aware value that the greedy allocation uses.

\begin{table}[htb]
\centering
\caption{Simple PCA and our full method with predicted coefficients on the benchmarks of Table~\ref{tab:main}, at the memory budgets of Table~\ref{tab:instantiation}. FlexAvatar is evaluated in self-reenactment on VFHQ, with the enrollment for both models. \textbf{Best} results. $^\ast$With respect to the host model. $^\dagger$Cross-reenactment. The higher ID and CSIM of simple PCA reflect under-animation, since a static head obtains a CSIM of 0.813 on VFHQ.}
\label{tab:supp_simplepca}
\scriptsize
\setlength{\tabcolsep}{1.6pt}
\begin{tabular}{@{}l ccccc ccccc cccc@{}}
\toprule
 & \multicolumn{5}{c}{AGORA (FFHQ)} & \multicolumn{5}{c}{FlexAvatar (VFHQ)} & \multicolumn{4}{c}{DynaAvatar (4D-Dress)} \\
\cmidrule(lr){2-6}\cmidrule(lr){7-11}\cmidrule(lr){12-15}
 & PSNR$^\ast\uparrow$ & SSIM$^\ast\uparrow$ & FID$\downarrow$ & ID$\uparrow$ & AED$\downarrow$ & PSNR$\uparrow$ & SSIM$\uparrow$ & LPIPS$\downarrow$ & CSIM$^\dagger\uparrow$ & AED$^\dagger\downarrow$ & PSNR$\uparrow$ & SSIM$\uparrow$ & LPIPS$\downarrow$ & PSNR$^\ast\uparrow$ \\
\midrule
Simple PCA  & 28.82 & 0.894 & 3.50 & \textbf{0.797} & 0.792 & 21.41 & 0.799 & 0.119 & \textbf{0.75} & 0.271 & 22.59 & 0.956 & 0.069 & 21.27 \\
GALA (ours) & \textbf{31.18} & \textbf{0.929} & \textbf{3.47} & 0.761 & \textbf{0.705} & \textbf{22.31} & \textbf{0.818} & \textbf{0.109} & 0.68 & \textbf{0.245} & \textbf{23.22} & \textbf{0.957} & \textbf{0.066} & \textbf{23.87} \\
\bottomrule
\end{tabular}
\end{table}

\section{Comparison to Alternative Distillation}
\label{sec:supp_ugb}

\paragraph{Distilled networks.}
The distilled networks of Figure~\ref{fig:pareto} follow the standard recipe of knowledge distillation~\citep{bucilua2006compression,hinton2015distilling}: a student network is trained with a mean-squared loss on the outputs of the frozen host model, here normalized per attribute group, on the same samples as our bases and coefficient networks. Every student is a reduced copy of the animation network of its host model, and all other parts of the host model are kept frozen. For AGORA, the student is the convolutional deformation branch with fewer channels, initialized from scratch or, for A5, by magnitude pruning of the channels of the host~\citep{pruningfilters}. For FlexAvatar and DynaAvatar, the student keeps 1 to 8 of the blocks of the transformer that is evaluated at every frame, initialized from the host and reduced in width. The DynaAvatar students additionally attend over a farthest-point-sampled subset of the Gaussians and interpolate the predicted residual to all 40{,}000 Gaussians. Table~\ref{tab:distilled} lists the size, the CPU time and the quality of every network.

\begin{table}[htb]
\centering
\caption{Distilled networks of Figure~\ref{fig:pareto}. For AGORA, channels are listed per resolution of the deformation branch. $L$ and $d$ are the number and the width of the transformer blocks, and $N$ the number of attended Gaussians for DynaAvatar. Size is the storage of the student in single precision. CPU times are for the animation step at batch size one with eight threads on an Intel Core Ultra 9 285K, without the skinning of DynaAvatar. Quality is the vertical axis of Figure~\ref{fig:pareto}: FID on FFHQ for AGORA and LPIPS with respect to the ground truth for FlexAvatar and DynaAvatar.}
\label{tab:distilled}
\footnotesize
\setlength{\tabcolsep}{3.5pt}
\begin{tabular}{@{}l l r r r r@{}}
\toprule
 & student & parameters & size & CPU & FID / LPIPS $\downarrow$ \\
\midrule
\multicolumn{6}{@{}l}{\emph{AGORA}, deformation branch with 512 / 256 / 128 channels at $64^2$ / $128^2$ / $256^2$} \\
A1 & 64 / 32 / 16 channels & 0.56~M & 2.2~MB & 9.4~ms & 4.14 \\
A2 & 192 / 96 channels, no $256^2$ block & 1.03~M & 4.1~MB & 12.2~ms & 3.73 \\
A3 & 128 / 64 / 32 channels & 0.89~M & 3.6~MB & 16.2~ms & 3.71 \\
A4 & 256 / 128 / 64 channels & 1.77~M & 7.1~MB & 45.1~ms & 3.71 \\
A5 & 448 / 224 / 112 channels, pruned from the host & 3.60~M & 14.4~MB & 124.4~ms & 3.60 \\
\midrule
\multicolumn{6}{@{}l}{\emph{FlexAvatar}, expression transformer with $L=8$, $d=768$} \\
F1 & $L=2$, $d=192$, upsampler of the host at 3/8 width & 4.6~M & 18.5~MB & 23.3~ms & 0.167 \\
F2 & $L=1$, $d=576$ & 9.9~M & 39.4~MB & 26.7~ms & 0.131 \\
F3 & $L=2$, $d=768$ & 27.2~M & 108.7~MB & 49.6~ms & 0.119 \\
F4 & $L=8$, $d=768$ & 112.2~M & 448.8~MB & 175.5~ms & 0.100 \\
\midrule
\multicolumn{6}{@{}l}{\emph{DynaAvatar}, dynamic transformer with $L=5$, $d=1{,}024$, $N=40{,}000$} \\
D1 & $L=1$, $d=64$, $N=252$ & 1.48~M & 5.9~MB & 3.9~ms & 0.0702 \\
D2 & $L=2$, $d=320$, $N=752$ & 11.6~M & 46.3~MB & 12.9~ms & 0.0675 \\
D3 & $L=2$, $d=512$, $N=2{,}500$ & 21.9~M & 87.4~MB & 82.2~ms & 0.0666 \\
\bottomrule
\end{tabular}
\end{table}

\paragraph{Results.}
Figure~\ref{fig:pareto} compares these networks with our method for a range of memory budgets. For all three host models, none of the distilled networks is both faster and more accurate than any of our models, and among the models that run in less than 80~ms the most accurate one is ours. For AGORA and DynaAvatar, none of the distilled networks reaches the quality of our model of Table~\ref{tab:main}. The best distilled AGORA network obtains an FID of 3.60 in 124~ms, whereas our model obtains 3.47 in 5.5~ms. For FlexAvatar, only the largest transformer approaches the LPIPS of the host model. It requires 175~ms, which is 40 times slower than our model. All distilled networks faster than 50~ms are less accurate than our smallest model, which has a 12~MB basis and takes 2~ms. The time of our method also grows slowly with memory, \eg a 17 times larger AGORA basis requires 5 times more time, which we attribute mainly to the memory bandwidth of the CPU. Finally, the difference between our models with predicted and with projected coefficients shows the effect of coefficient prediction on the final quality. For larger budgets on FlexAvatar and DynaAvatar, projected coefficients bring our models close to the host model, which indicates that the remaining error is mostly due to coefficient prediction.

\section{Comparison to Per-Subject Blendshape Avatars}
\label{sec:supp_per_subject}

Per-subject methods fit an avatar to a video or a multi-view capture of each person, whereas our model initializes from one photo through the host model and reuses one shared basis.

\begin{figure}[htbp]
\centering
\includegraphics[width=\linewidth]{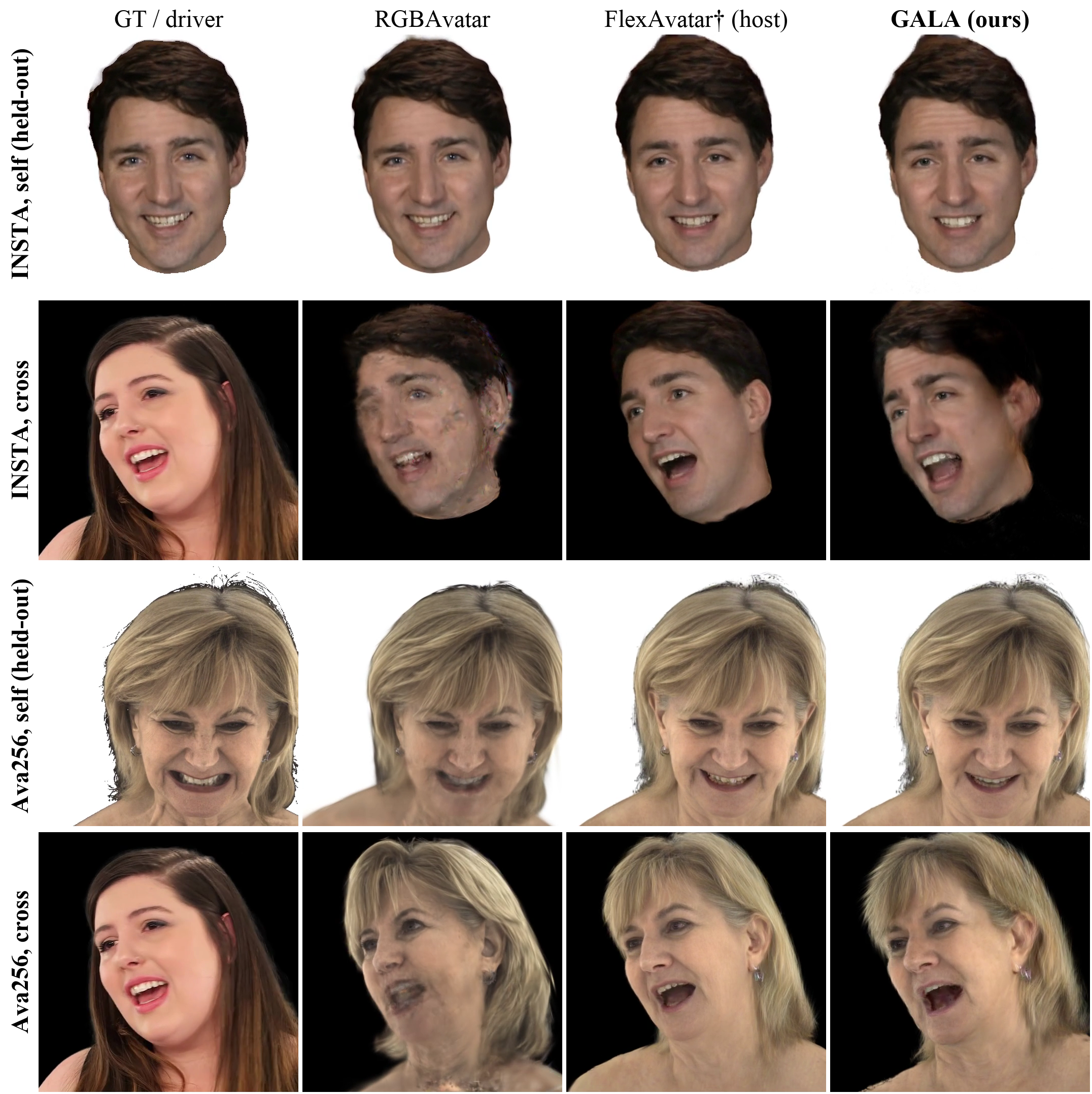}
\caption{Comparison to RGBAvatar on INSTA and Ava256. RGBAvatar is fitted to the video (INSTA) or to the 15-view capture (Ava256) of the subject, whereas the host model FlexAvatar and GALA use one photo. The self rows show held-out frames next to the ground truth, the cross rows are driven by VFHQ videos of other people (left) without ground truth.}
\label{fig:per_subject_rgbavatar}
\end{figure}

\begin{figure}[htbp]
\centering
\includegraphics[width=0.85\linewidth]{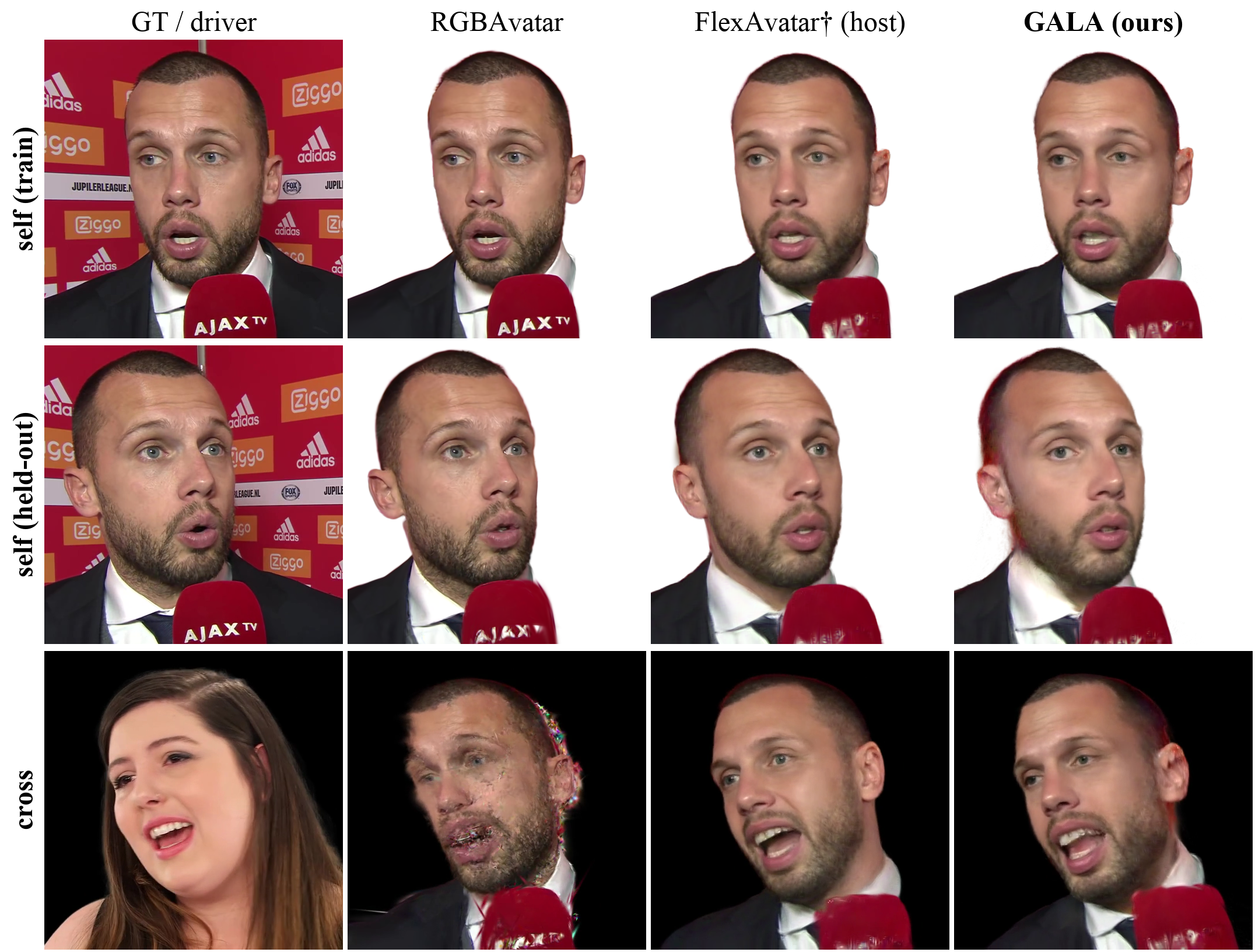}
\caption{Comparison to RGBAvatar on a VFHQ clip of Table~\ref{tab:per_subject_vfhq}. RGBAvatar is fitted to the first 70\% of the clip, whereas the host model FlexAvatar and GALA use its first frame. Rows: a training frame of RGBAvatar, a held-out frame, and cross reenactment driven by a VFHQ video of another person (left).}
\label{fig:per_subject_vfhq}
\end{figure}

\begin{figure}[htbp]
\centering
\includegraphics[width=0.85\linewidth]{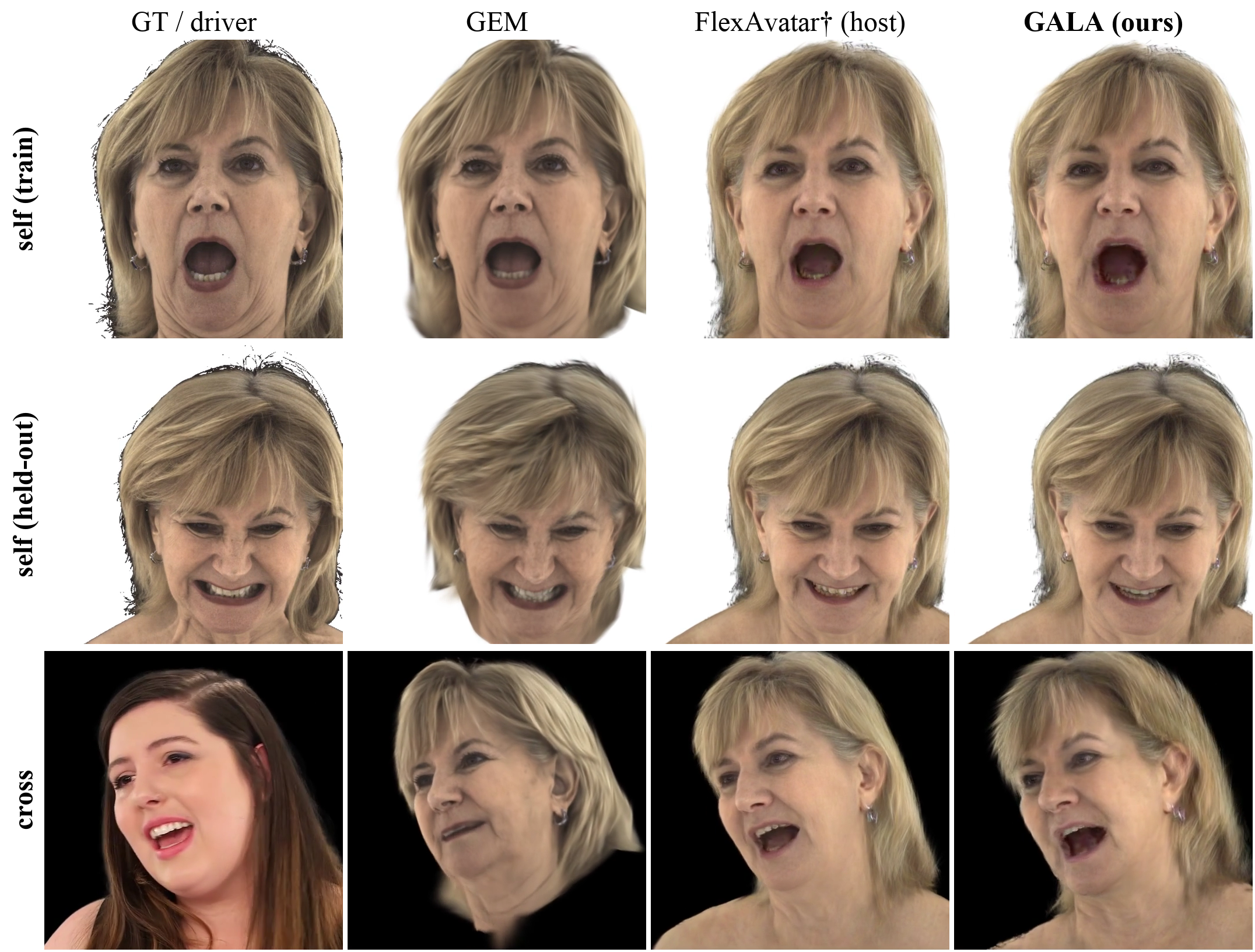}
\caption{Comparison to GEM on an Ava256 identity. GEM is fitted to the 15-view capture of the subject, whereas the host model FlexAvatar and GALA use one photo. Rows: a training frame of GEM from one of its training cameras, a held-out frame from the camera of the input photo, and cross reenactment driven by a VFHQ video of another person (left).}
\label{fig:per_subject_gem}
\end{figure}

\paragraph{Evaluation details.} Table~\ref{tab:per_subject} compares FlexAvatar + GALA with RGBAvatar~\citep{rgbavatar} on 6 held-out identities of INSTA~\citep{insta}, and with GEM~\citep{zielonka2025gem} and RGBAvatar on 3 identities of Ava256~\citep{ava256}. RGBAvatar is fitted to the monocular video of each INSTA subject and, like GEM, to the 15-view capture of each Ava256 subject. Our model initializes from one photo of the subject. Self reenactment is scored on training frames of the per-subject fits (the first 250 frames of each INSTA video, and 120 Ava256 tuples per identity from 4 training cameras, none of them the camera of the input photo) and on held-out frames (the last 350 frames of each INSTA video, and 1{,}384 Ava256 tuples with unseen motion and mostly unseen views), cross reenactment on 3 VFHQ drivers without ground truth. Both per-subject fits are driven with their released pipelines: RGBAvatar with the drivers tracked by the same tracker and configuration as its training frames, and GEM with its image-based coefficient regressor relative to a neutral reference frame of each driver. Table~\ref{tab:per_subject_vfhq} repeats the comparison with RGBAvatar on 3 short clips of the VFHQ test set~\citep{VFHQ}, where RGBAvatar is fitted to the first 70\% of each clip (127 frames on average) with the tracking of its released pipeline, FlexAvatar and our model use the first frame, and self reenactment is scored on the last 20\% of each clip, with PSNR and LPIPS on 10 frames per clip. On 4D-Dress~\citep{dress4d}, WebAvatar~\citep{webavatar} is fitted to 24 views of each subject (999 frames per view, 4.8~h) and scored on a held-out take. Figures~\ref{fig:per_subject_rgbavatar} to \ref{fig:per_subject_gem} show frames.

\paragraph{Results.} In self reenactment, the per-subject fits score better: they have a lower AED on the held-out frames of every dataset, and RGBAvatar a higher PSNR on INSTA and VFHQ. This is expected. Each per-subject fit minimizes a photometric loss on the very sequence that self reenactment scores, from one video (INSTA, VFHQ) or from 15 views of the capture (Ava256), whereas our model never sees this sequence and is not optimized for its reconstruction. Our higher held-out PSNR on Ava256 comes from the camera of the input photo, on which the per-subject fits were not trained. The intended use of an avatar is cross reenactment, animation by the motion of another person, and there our model is ahead: it has a lower AED and APD than every per-subject fit on all three datasets and a higher CSIM on INSTA and VFHQ, whereas GEM keeps a higher CSIM on Ava256. Figures~\ref{fig:per_subject_rgbavatar} to \ref{fig:per_subject_gem} show why: on their own held-out frames the per-subject fits are close to the ground truth, but under the motion of the driver RGBAvatar breaks up at head poses absent from its training video and GEM barely opens the mouth, whereas our model follows the driver. On 4D-Dress, WebAvatar obtains 22.78~dB, compared with 21.82~dB for DynaAvatar + GALA and 22.93~dB for DynaAvatar. On their own training frames, RGBAvatar reaches 33.21~dB on INSTA and 23.35~dB on Ava256, whereas GEM reaches 16.64~dB. Two properties of GEM explain its numbers. GEM does not model the neck and shoulders, which its pipeline masks out of its training images and which the ground truth keeps: this region covers 3\% of the pixels but 45\% of GEM's squared error, and GEM's own training image, scored as a render, reaches 19.3~dB. Its fit is also blurry from its first stage, which reaches 21.2~dB against its own neck-free training images before the basis and the coefficient regressor are fitted. On the head region alone, GEM reaches 20.05~dB on its training frames and RGBAvatar 23.92~dB.

\begin{table}[htbp]
\centering
\caption{Comparison to per-subject blendshape avatars on 6 held-out identities of INSTA and 3 identities of Ava256. Input per subject: views, frames per view and time of the fit or of our enrollment, without face tracking. CSIM uses the input photo as reference. \textbf{Best} held-out and cross results, excluding the host model ($^\dagger$).}
\label{tab:per_subject}
\footnotesize
\setlength{\tabcolsep}{2.5pt}
\begin{tabular}{@{}l ccc cc ccc ccc@{}}
\toprule
 & \multicolumn{3}{c}{Input per subject} & \multicolumn{5}{c}{Self reenactment} & \multicolumn{3}{c}{Cross reenactment} \\
\cmidrule(lr){2-4}\cmidrule(lr){5-9}\cmidrule(lr){10-12}
 & & & & \multicolumn{2}{c}{training frames} & \multicolumn{3}{c}{held-out frames} & & & \\
\cmidrule(lr){5-6}\cmidrule(lr){7-9}
 & Views & Frames & Fit & PSNR$\uparrow$ & LPIPS$\downarrow$ & PSNR$\uparrow$ & LPIPS$\downarrow$ & AED$\downarrow$ & CSIM$\uparrow$ & AED$\downarrow$ & APD$\downarrow$ \\
\midrule
\multicolumn{12}{@{}l}{\emph{INSTA}} \\
RGBAvatar & 1 & 2{,}790 & 55\,s & 33.21 & 0.0487 & \textbf{30.23} & \textbf{0.0518} & \textbf{0.173} & 0.524 & 0.898 & 0.0499 \\
\gr{FlexAvatar$^\dagger$} & \gr{1} & \gr{1} & \gr{$\sim$1\,min} & \gr{24.58} & \gr{0.0858} & \gr{24.43} & \gr{0.0675} & \gr{0.311} & \gr{0.612} & \gr{0.700} & \gr{0.0219} \\
GALA (ours) & 1 & 1 & $\sim$1\,min & 23.76 & 0.1152 & 23.80 & 0.0812 & 0.387 & \textbf{0.639} & \textbf{0.800} & \textbf{0.0234} \\
\midrule
\multicolumn{12}{@{}l}{\emph{Ava256}} \\
GEM & 15 & 2{,}921 & 27.8\,h & 16.64 & 0.2792 & 15.37 & 0.2949 & \textbf{0.457} & \textbf{0.674} & 1.006 & 0.0505 \\
RGBAvatar & 15 & 2{,}921 & 32.7\,min & 23.35 & 0.2076 & 18.12 & 0.2489 & \textbf{0.457} & 0.549 & 0.914 & 0.0385 \\
\gr{FlexAvatar$^\dagger$} & \gr{1} & \gr{1} & \gr{$\sim$1\,min} & \gr{16.42} & \gr{0.2974} & \gr{19.22} & \gr{0.2082} & \gr{0.512} & \gr{0.563} & \gr{0.736} & \gr{0.0245} \\
GALA (ours) & 1 & 1 & $\sim$1\,min & 16.49 & 0.2985 & \textbf{18.81} & \textbf{0.2153} & 0.579 & 0.590 & \textbf{0.852} & \textbf{0.0247} \\
\bottomrule
\end{tabular}
\end{table}

\begin{table}[htbp]
\centering
\caption{Comparison to RGBAvatar on 3 short clips of the VFHQ test set. RGBAvatar is fitted to the first 70\% of each clip, FlexAvatar and our model use the first frame, and self reenactment is scored on the last 20\%. \textbf{Best} results, excluding the host model ($^\dagger$).}
\label{tab:per_subject_vfhq}
\footnotesize
\setlength{\tabcolsep}{2.5pt}
\begin{tabular}{@{}l ccc ccc ccc@{}}
\toprule
 & \multicolumn{3}{c}{Input per subject} & \multicolumn{3}{c}{Self reenactment, held-out frames} & \multicolumn{3}{c}{Cross reenactment} \\
\cmidrule(lr){2-4}\cmidrule(lr){5-7}\cmidrule(lr){8-10}
 & Views & Frames & Fit & PSNR$\uparrow$ & LPIPS$\downarrow$ & AED$\downarrow$ & CSIM$\uparrow$ & AED$\downarrow$ & APD$\downarrow$ \\
\midrule
RGBAvatar & 1 & 127 & 65\,s & \textbf{20.70} & \textbf{0.1084} & \textbf{0.251} & 0.451 & 0.860 & 0.0645 \\
\gr{FlexAvatar$^\dagger$} & \gr{1} & \gr{1} & \gr{$\sim$1\,min} & \gr{19.72} & \gr{0.1363} & \gr{0.292} & \gr{0.628} & \gr{0.634} & \gr{0.0188} \\
GALA (ours) & 1 & 1 & $\sim$1\,min & 19.15 & 0.1496 & 0.325 & \textbf{0.666} & \textbf{0.787} & \textbf{0.0212} \\
\bottomrule
\end{tabular}
\end{table}

\clearpage
\section{Additional Qualitative Results}
\label{sec:supp_qualitative}

\begin{figure}[htbp]
    \centering
    \hspace*{\qlabw}\qh{6}{Host model}\qh{6}{Neutral only}\qh{6}{Projected}%
    \qh{6}{GALA (ours)}\qh{6}{Error, projected}\qh{6}{Error, GALA}\par\vspace{1pt}
    \qrow{AGORA}{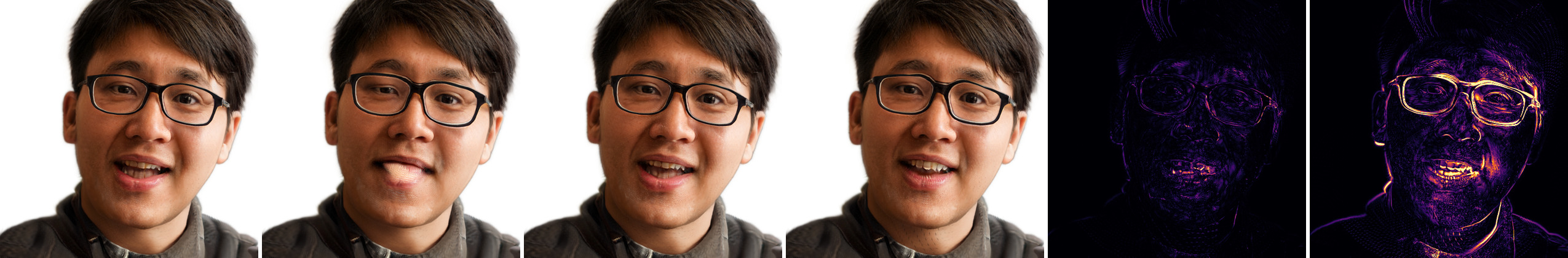}
    \qrow{AGORA}{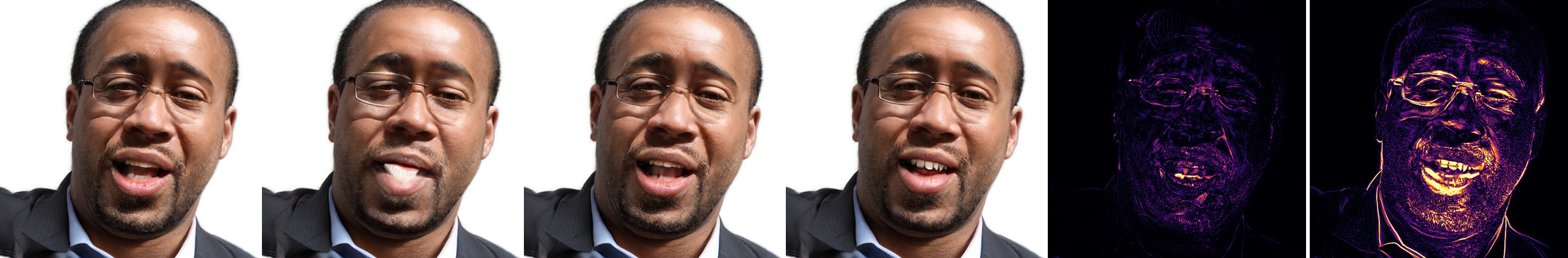}
    \qrow{FlexAvatar}{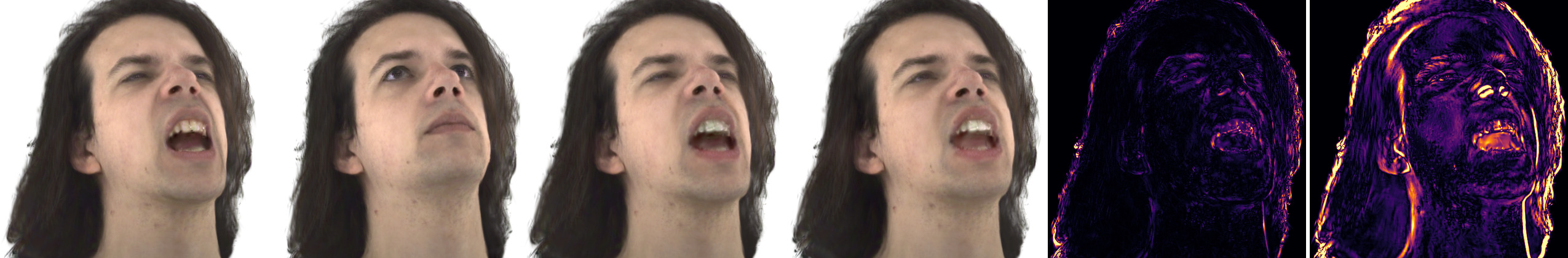}
    \qrow{FlexAvatar}{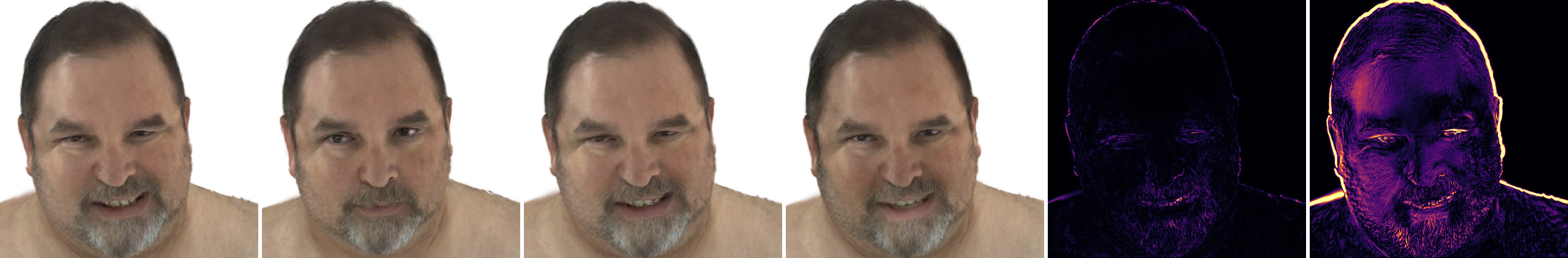}
    \qrow{DynaAvatar}{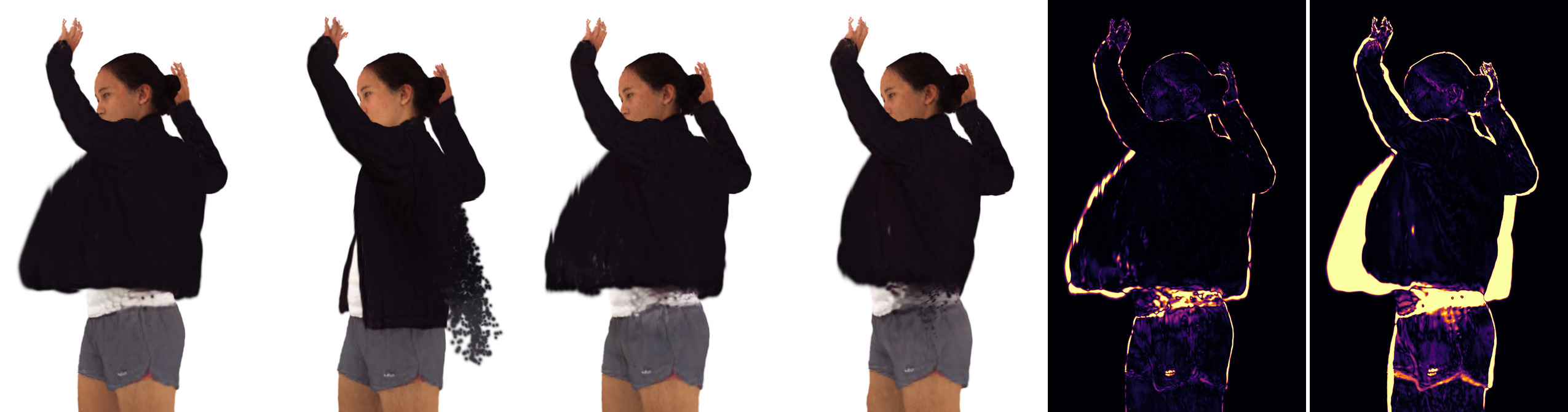}
    \caption{More difficult examples, with the error scale of Figure~\ref{fig:qualitative}. \emph{Neutral only} is the neutral avatar with all coefficients zero, \emph{Projected} uses projected coefficients $c^\star$ and \emph{GALA} the predicted ones. Rows 1 and 2: held-out AGORA identities at the frame with the largest residual of the host model. Rows 3 and 4: held-out Ava256 subjects in the few-shot and the single-shot setting. Row 5: the frame with the largest error of our model in the DynaAvatar sequence of Figure~\ref{fig:qualitative}. In row 4 the coefficient network fails: 23.6~dB PSNR with respect to the host model with predicted coefficients, 40.2~dB with projected ones.}
    \label{fig:supp_qualitative}
\end{figure}

Figure~\ref{fig:supp_qualitative} shows more difficult examples in the layout of Figure~\ref{fig:qualitative}, including a subject for which the coefficient network fails.

\section{Visualization of Blendshapes}
\label{sec:supp_blendshapes}

\begin{figure}[htbp]
    \centering
    \hspace*{\qlabw}\qh{7}{Where}\qhh{7}{Identity 1}\qhh{7}{Identity 2}%
    \qhh{7}{Identity 3}\par\vspace{1pt}
    \qrow{AGORA}{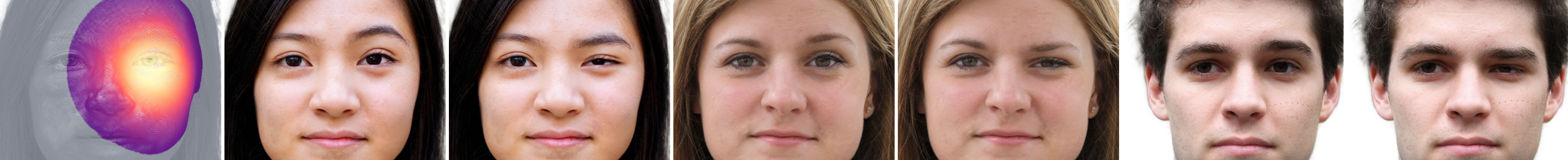}
    \qrow{FlexAv.}{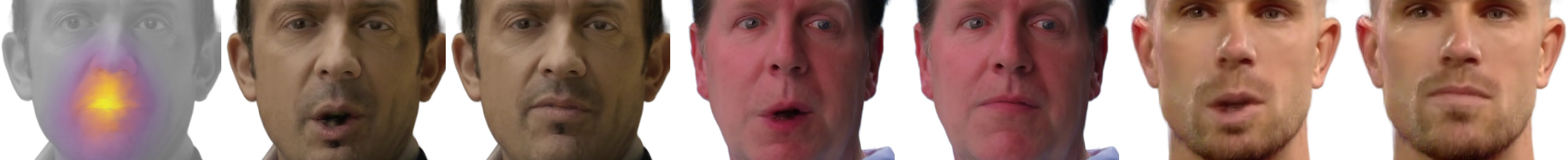}
    \qrow{DynaAvatar}{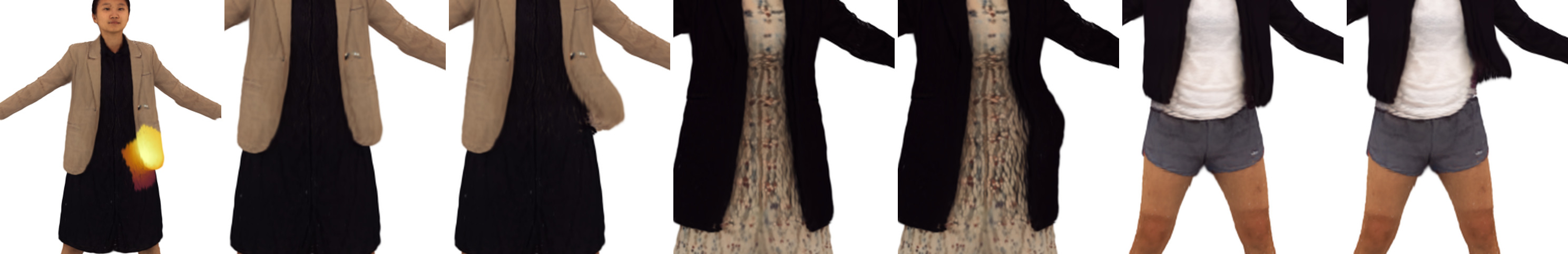}
    \caption{Local motions learned by our blendshapes. The first column highlights the region affected by the visualized blendshapes, and the other columns show two states of their motion for three identities: closing of an eye (AGORA), closing of the mouth (FlexAvatar) and lifting of the front part of an open garment (DynaAvatar).}
    \label{fig:blendshapes}
\end{figure}

Figure~\ref{fig:blendshapes} visualizes motions captured by our basis. Each local basis $U_{a,j}$ is supported on one spatial block (Section~\ref{subsubsec:improvedpca}), hence our blendshapes learn local motions that correspond to particular regions of the avatar, and the same blendshapes produce consistent motions for different identities. As examples we show blendshapes that control an eye for AGORA, the mouth for FlexAvatar, and a moving part of the garment for DynaAvatar, which is shown in the canonical A-pose.

\paragraph{Construction.} For the head models, a Gaussian window is placed on a facial region of the shared template, and the components with at least a quarter of their displacement energy inside the window form a group (737 and 945 components for the AGORA rows, 349 and 343 for the FlexAvatar rows). The group is driven along the leading eigenvector of the covariance of its coefficients over the training data, and the two states are the means of the coefficient vectors near the 2nd and the 98th percentile along this direction, applied through the window, with the amplitude for FlexAvatar limited to 14~mm for a head of 230~mm. For DynaAvatar, the first state is the neutral avatar, and the second sets one component to its 2nd or 98th percentile over the training data and the other coefficients to the mean of the nearest 10\% of the coefficient vectors, attenuated toward the neutral avatar away from the component. The rows therefore show conditional activations of parts of the basis, not isolated basis vectors. The first column shows the window for the head models and the displacement field of the component for DynaAvatar, normalized per panel. The AGORA and FlexAvatar identities are held out, the DynaAvatar subjects belong to the training data. Bases: 154.1, 48 and 88.3~MB.

\begin{figure}[htbp]
    \centering
    \hspace*{\qlabw}\qh{7}{Where}\qhh{7}{Identity 1}\qhh{7}{Identity 2}%
    \qhh{7}{Identity 3}\par\vspace{1pt}
    \qrow{AGORA}{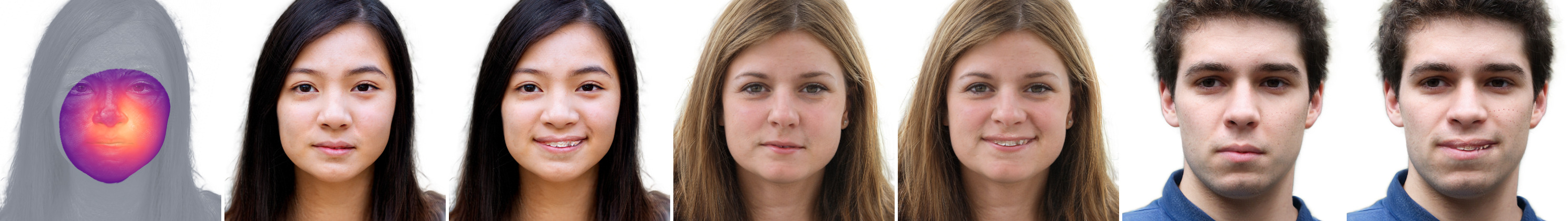}
    \qrow{FlexAvatar}{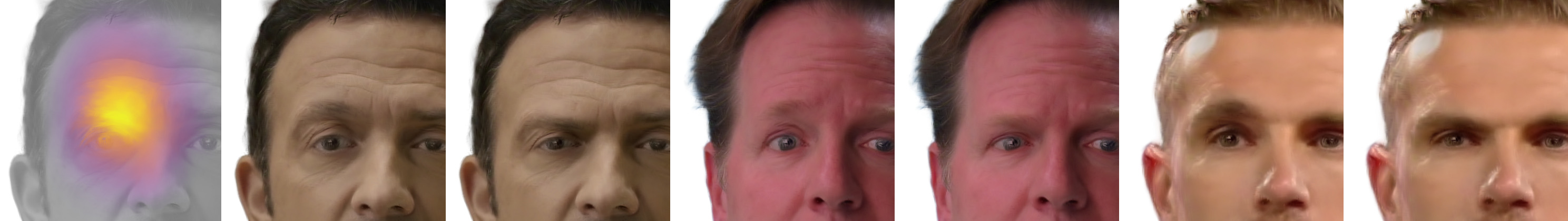}
    \qrow{DynaAvatar}{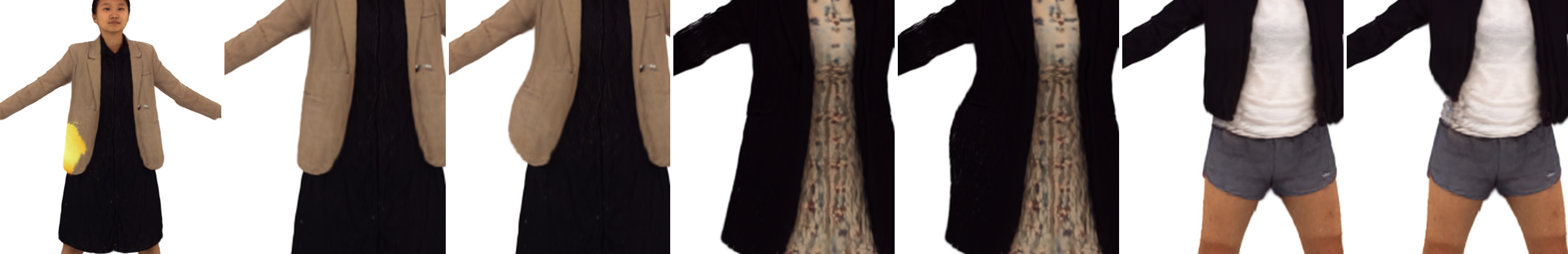}
    \caption{Further local motions, in the layout of Figure~\ref{fig:blendshapes}: a smile (AGORA), lowering of a brow together with the upper eyelid (FlexAvatar, cropped to the brow) and the opposite front part of the garment (DynaAvatar).}
    \label{fig:supp_blendshapes}
\end{figure}